\documentclass[lettersize,journal]{IEEEtran}
\usepackage{amsmath,amsfonts}
\usepackage{algorithmic}
\usepackage{algorithm}
\usepackage{array}
\usepackage[caption=false,font=normalsize,labelfont=sf,textfont=sf]{subfig}
\usepackage{textcomp}
\usepackage{stfloats}
\usepackage{url}
\usepackage{verbatim}
\usepackage{graphicx}
\usepackage{cite}
\usepackage{booktabs}  %% 三线表
\usepackage{diagbox}   %% 斜线表头
\usepackage{multirow}  %% 合并单元格
\begin{document}

\title{Multi-Scale Spatial-Temporal Recurrent Networks\\for Traffic Flow Prediction}

\author{Haiyang Liu, Chunjiang Zhu, Detian Zhang, and Qing Li,~\IEEEmembership{Fellow,~IEEE}
        % <-this % stops a space
\thanks{Haiyang Liu and Detian Zhang are with the Institute of Artificial Intelligence, Department of Computer Science and Technology, Soochow University, Suzhou, China. (e-mail: 20215227052@stu.suda.edu.cn; detian@suda.edu.cn). Chunjiang Zhu is with the Department of Computer Science, University of North Carolina at Greensboro, Greensboro, NC, USA. (e-mail: chunjiang.zhu@uncg.edu). Qing Li is with the Department of Computing, The Hong Kong Polytechnic University, Hong Kong, China. (e-mail: qing-prof.li@polyu.edu.hk).}% <-this % stops a space
\thanks{Chunjiang Zhu is supported by UNCG Start-up Funds and Faculty First Award. Detian Zhang is partially supported by the Collaborative Innovation Center of Novel Software Technology and Industrialization, the Priority Academic Program Development of Jiangsu Higher Education Institutions.}
\thanks{Corresponding author: Detian Zhang.}}

% The paper headers
\markboth{Journal of \LaTeX\ Class Files,~Vol.~14, No.~8, August~2021}%
{Shell \MakeLowercase{\textit{et al.}}: A Sample Article Using IEEEtran.cls for IEEE Journals}

\IEEEpubid{0000--0000/00\$00.00~\copyright~2021 IEEE}
% Remember, if you use this you must call \IEEEpubidadjcol in the second
% column for its text to clear the IEEEpubid mark.

\maketitle

\begin{abstract}
Traffic flow prediction is one of the most fundamental tasks of intelligent transportation systems. The complex and dynamic spatial-temporal dependencies make the traffic flow prediction quite challenging. Although existing spatial-temporal graph neural networks hold prominent, they often encounter challenges such as  (1) ignoring the fixed graph that limits the predictive performance of the model, (2) insufficiently capturing complex spatial-temporal dependencies simultaneously, and (3) lacking attention to spatial-temporal information at different time lengths. In this paper, we propose a Multi-Scale Spatial-Temporal Recurrent Network for traffic flow prediction, namely MSSTRN, which consists of two different recurrent neural networks: the single-step gate recurrent unit and the multi-step gate recurrent unit to fully capture the complex spatial-temporal information in the traffic data under different time windows. Moreover, we propose a spatial-temporal synchronous attention mechanism that integrates adaptive position graph convolutions into the self-attention mechanism to achieve synchronous capture of spatial-temporal dependencies. We conducted extensive experiments on four real traffic datasets and demonstrated that our model achieves the best prediction accuracy with non-trivial margins compared to all the twenty baseline methods.
\end{abstract}


\section{Introduction}
\IEEEPARstart{I}{n} order to effectively relieve the pressure on urban transportation facilities caused by the increasing population and vehicles, more and more cities have begun to vigorously develop Intelligent Transportation Systems (ITS) \cite{zhang2011data,veres2019deep}. As one of the main functions of ITS, traffic flow prediction can effectively alleviate traffic congestion, reduce traffic costs, and improve traffic management efficiency \cite{lv2014traffic,zheng2014urban,wu2020comprehensive,jiang2021dl}.However, traffic flow prediction remains a challenging task due to the intricate spatial-temporal dependencies inherent in traffic data.

Initially, traffic flow prediction was regarded as a time series forecasting task, with researchers primarily focusing on non-deep learning techniques \cite{drucker1996support,lee1999application,williams2003modeling,wu2004travel,zivot2006vector,van2012short,zheng2014short,cai2016spatiotemporal}. However, these methods were limited in their ability to capture only linear relationships among spatial nodes. 
With the continuous improvement of computer processing power, deep learning methods have found widespread applications. Recurrent Neural Networks (RNNs) \cite{LSTM1997long,GRU2014properties} and Convolutional Neural Networks (CNNs) have proven effective in capturing the nonlinear spatial-temporal dependencies in structured traffic data. Graph Neural Networks (GNNs) have gained popularity for spatial modeling in traffic prediction, thanks to their ability to capture spatial relationships in irregular and unstructured spatial-temporal data.
Then Spatial-Temporal Graph Neural Networks (STGNNs) have emerged as frameworks that combine GNNs with temporal modules (e.g., RNNs \cite{sutskever2014sequence,li2018dcrnn_traffic,bai2020adaptive,chen2021z,yu2022regularized,liu2022msdr}, CNNs \cite{yu2018spatio,wu2019graph,huang2020lsgcn,song2020spatial,li2021spatial,zhao2022spatial} or the attention mechanism \cite{guo2019attention,wang2020traffic,zheng2020gman,lan2022dstagnn,cirstea2022towards,liu2023attention}) to model spatial-temporal dependencies. These models have gained significant traction in the field of traffic prediction and have showcased remarkable predictive performance when compared with earlier statistical models and traditional machine learning methods.

\begin{figure}[!t]
\centering
\includegraphics[width=3.0in]{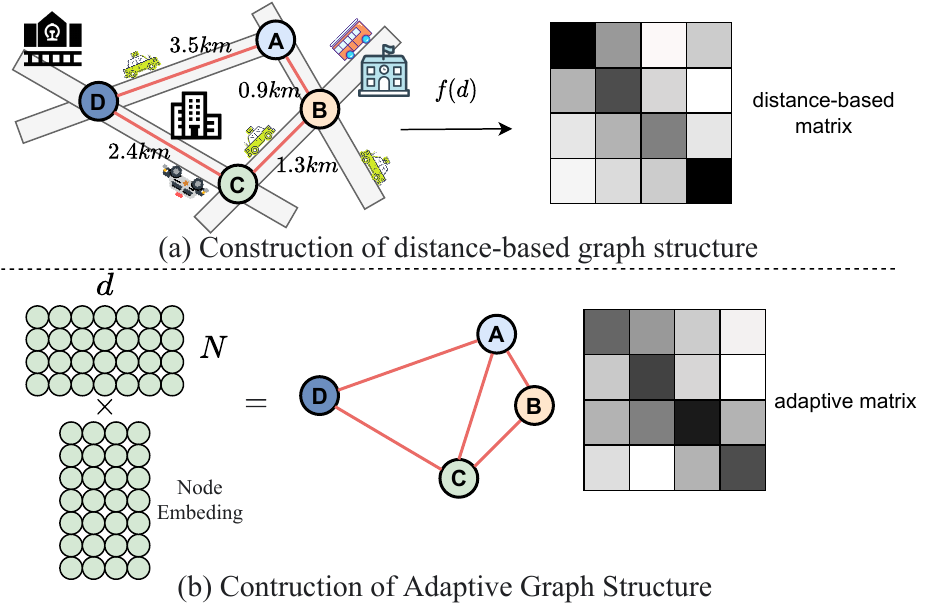} % Reduce the figure size so that it is slightly narrower than the column.
% \vspace{-0.1in}
\caption{Two common graph construction methods.}
\label{Example1}
\end{figure}

\IEEEpubidadjcol

\begin{figure}[!t]
\centering
\includegraphics[width=3.2in]{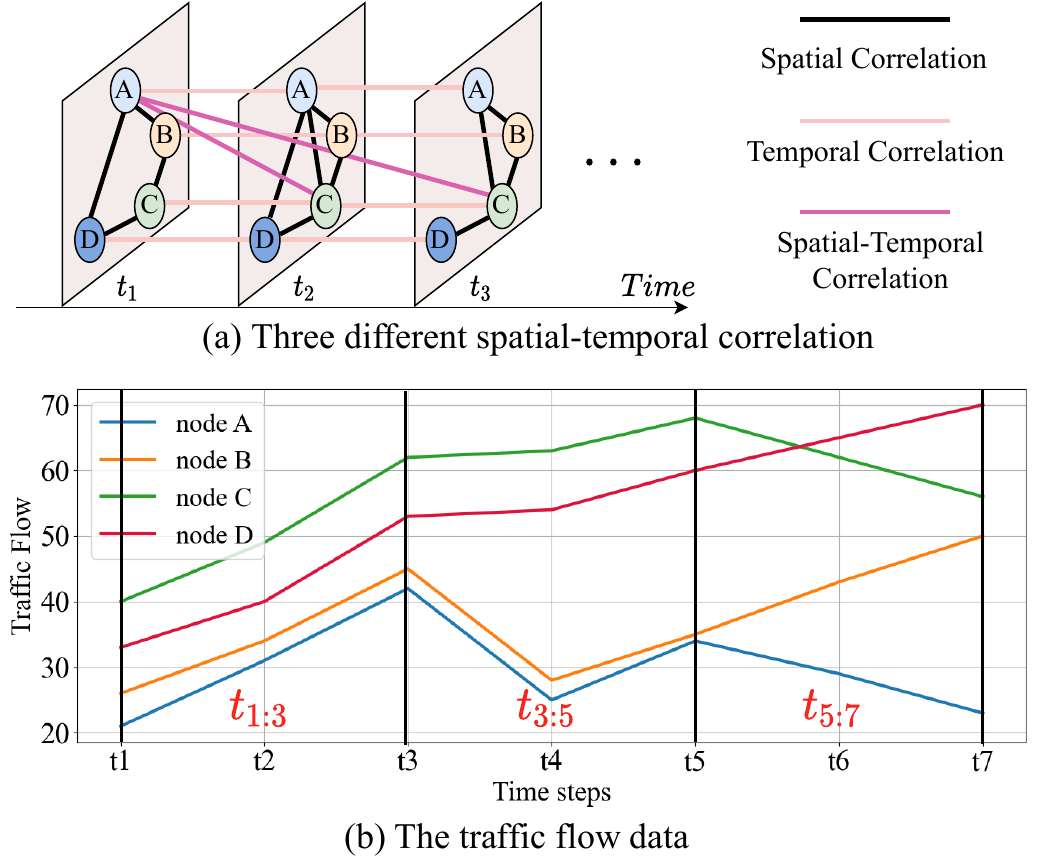} % Reduce the figure size so that it is slightly narrower than the column.
% \vspace{-0.1in}
\caption{An example of visualizing the spatial-temporal relationship of traffic flow. $A$, $B$, $C$, $D$ four nodes influence each other and flow trend.}
\label{Example2}
\end{figure}

Despite the remarkable performance exhibited by current STGNNs in traffic forecasting, there are still three crucial issues that have either been overlooked or not fully addressed.
\begin{itemize}
\item Firstly, the predictive performance of the models is constrained by fixed graphs. Although there exists research on spatial modeling that employs static graphs constructed using distance (Fig. \ref{Example1}(a)) \cite{yu2018spatio,guo2021learning,zhang2022robust} or initialised node embeddings to construct learnable adaptive graphs (Fig. \ref{Example1}(b)) \cite{wu2019graph,bai2020adaptive,chen2021z,yu2022regularized}, the graph structure used for spatial modeling remains uniform across all time steps, with constant relationship weights between nodes. However, the impact and relationships between nodes may vary across different time positions.
\item Secondly, there is a disregard for the simultaneous modeling of spatial-temporal correlations. In Fig. \ref{Example2}(a), the information in traffic data can be categorized into three aspects: spatial correlation, temporal correlation and spatial-temporal correlation. Many existing approaches, such as STGCN \cite{yu2018spatio} and Graph WaveNet \cite{wu2019graph}, achieve spatial-temporal modeling through separate temporal and spatial modeling, which can be considered as asynchronous spatial-temporal modeling. While some recent works \cite{song2020spatial,li2021spatial} have made progress in capturing simultaneous spatial-temporal dependencies by constructing spatial-temporal graphs, their use of static spatial-temporal graphs inherently limits their effectiveness. Additionally, expanding the perceptual field by adding stacked layers can result in a significant increase in computational cost.
\item Thirdly, there is a lack of attention given to the spatial-temporal information between time steps of different sizes. In RNN-based methods, the time window $s$ of the input node is typically set to $1$, allowing for the update of current time step information based on historical time step information. On the other hand, CNN-based or attention-based methods set $s$ to the entire historical time step $T$, enabling the analysis of spatial-temporal relationships across all time steps. However, both approaches fail to explore the spatial-temporal relationship between different sub-time windows. 

As illustrated in Fig. \ref{Example2}(b), nodes A, B, C, and D at time steps $t_1 \to t_2 \to t_3$ exhibit similar flow characteristics. A and B have similar flow trends, as do C and D at time steps $t_3 \to t_4 \to t_5$. Similarly, A and C have similar flow trends, while B and D have similar flow trends at time steps $t_5 \to t_6 \to t_7$. However, within the sub-time window ${t_{1:3}} \to {t_{3:5}}$ ($1<s=3<T$ for example), the flow trends of nodes A and B are similar, as are those of C and D. Yet, these four nodes exhibit different characteristics within ${t_{3:5}} \to {t_{5:7}}$. Although \cite{ye2022learning} tackles spatial(-only) modeling by constructing different graph structures for multiple time scales, it fails to address the modeling of diverse spatial-temporal dependencies at different time scales.
\end{itemize}
Therefore, we believe that the potential of STGNNs remains largely untapped, and addressing the aforementioned issues is of utmost importance in fully leveraging the complex spatial-temporal information present in traffic data to enhance model prediction performance.

To address these challenges, we introduce a novel deep learning model called Multi-Scale Spatial-Temporal Recurrent Networks (MSSTRN). Our proposed MSSTRN incorporates two distinct recurrent neural networks: the Single Step Gate Recurrent Unit (SS-GRU) and the Multi-Step Gate Recurrent Unit (MS-GRU). These networks integrate graph convolution and attention mechanisms into the GRU framework. The key innovation of our work lies in the simultaneous capture of spatial-temporal dependencies across different time windows. We summarize our main contributions as follows:

\begin{itemize}
\item Inspired by previous research on adaptive graph generation, we present a novel module Adaptive Position Graph Generation (APGG). This module utilizes node embeddings with learnable temporal position information at various scales to initialise adaptive graphs with different time windows and to continue learning during the training process.
\item For issue (2), we propose a Spatial-Temporal Synchronous Attention (STSAtt) that integrates Adaptive Position Graph Convolution Networks (APGCN) into self-attention to achieve the synchronous capture of spatial-temporal dependencies.
\item For issue (3), we integrate APGCN and STSAtt into the GRU architecture, resulting in two respective types of recurrent neural networks: SS-GRU and MS-GRU. These neural networks effectively capture the complex spatial-temporal information present in traffic data across different time windows.
\item We conducted extensive experiments using four real-world traffic flow datasets. The experimental results consistently demonstrate that our models outperform all baseline methods in terms of prediction performance. Furthermore, the model-building experiments indicate that different stacking approaches of SS-GRU and MS-GRU may lead to favorable prediction performance.
\end{itemize}

\section{Related Work}
\subsection{Traffic Flow Prediction}
Traffic prediction is one of the key functions of $ITS$, and related work has been carried out for many years with many excellent results. 

\textbf{Early non-deep learning methods} had difficulty capturing the spatial relationships in traffic data, and traffic flow forecasting was early on regarded as a time series forecasting problem. Statistical models such as Historical Average (HA), Vector Auto-Regressive (VAR) \cite{zivot2006vector}, Auto-Regressive Integrated Moving Average (ARIMA) \cite{lee1999application,williams2003modeling} were widely used. Although these methods are simple and fast to run, most of them can only consider the linear relationships of the nodes themselves and cannot handle the non-linear relationships in the traffic data, resulting in large forecast prediction errors.Flexible machine learning models had also attracted the attention of researchers, for example, K-Nearest Neighbour (KNN) \cite{van2012short,zheng2014short,cai2016spatiotemporal}, Support Vector Machines (SVR) \cite{drucker1996support,wu2004travel}. Although machine learning models can handle complex non-linear traffic data, their manual selection and extraction of features are not suitable for handling large-scale traffic data.

\textbf{Deep learning-based methods} are now widely used in traffic prediction problems due to their ability to effectively capture the temporal dependencies of nodes and the spatial dependencies between nodes. Traffic forecasting is essentially a time-series forecasting task with additional spatial attributes, so prior geographic knowledge has a huge impact on forecasting performance. CNN can obtain spatial information from grid data through convolution, but it cannot handle non-Euclidean data, and the use of grid data structures has its own limitations in representing spatial topology. At present, GNNs have become the basic modules of spatial modeling in traffic flow prediction research. As a type of GNNs, the emergence of Graph Convolutional Networks (GCNs) \cite{bruna2013spectral,atwood2016diffusion} enables deep learning models to handle non-Euclidean data and capture implicit spatial dependencies. It relies on defined non-Euclidean graphs, such as distance graphs \cite{li2018dcrnn_traffic,zheng2020gman,pan2019urban,park2020st,xu2020spatial,zhang2022robust}, binary graphs \cite{song2020spatial,huang2020lsgcn,guo2020dynamic}, and adaptive graphs \cite{wu2019graph,bai2020adaptive,chen2021z,yu2022regularized}, etc. Regardless of the spatial topology graph, the graph attention network (GAT) \cite{velivckovic2017graph} achieves spatial modeling with the help of attention mechanism \cite{guo2019attention,shi2020spatial,luo2023dynamic}, but the attention mechanism itself is quite time-consuming. EGS \cite{ye2022learning} proposes a recurrent construction approach, which uses \emph{GRU} to generate graph structures for each time node at different time scales. Spatial-Temporal Graph Neural Networks (STGNNs) are a commonly used class of traffic prediction frameworks that combine GNNs and temporal models (e.g., RNNs, TCNs or temporal attention) to model the spatial-temporal dependencies of traffic data. Thus, existing deep neural networks can be broadly classified into RNN-based approaches \cite{sutskever2014sequence,bai2020adaptive,chen2021z,yu2022regularized,liu2022msdr}, CNN-based approaches \cite{yu2018spatio,huang2020lsgcn,song2020spatial,li2021spatial}, and attention-based approaches \cite{lan2022dstagnn,cirstea2022towards} from the perspective of temporal modeling. 
RNN-based methods usually integrate spatial modules into recurrent bodies to construct spatial-temporal recurrent neural networks. DCRNN \cite{li2018dcrnn_traffic} uses diffusion graph convolution instead of fully connected layers in GRU to build sequence-to-sequence spatial-temporal models. CNN-based methods and attention-based methods are mostly stacked with spatial modules to capture spatial-temporal dependencies, and can expand the receptive field of the model by increasing the number of stacked layers. For example, Graph WaveNet \cite{wu2019graph} introduces an adaptive adjacency matrix and combines diffusion graph convolution with 1D convolution, and ASTGCN \cite{guo2019attention} combines spatial-temporal convolution and spatial-temporal attention mechanisms to effectively capture dynamic spatial-temporal features. In addition, differential equation-based methods \cite{fang2021spatial,choi2022graph} have also been used for spatial-temporal modeling of traffic data prediction with good results.

\subsection{Graph Convolutional Networks}
The advent of graph convolutional networks has enabled deep learning models to process non-Euclidean data, and it is commonly classified into two types: spectral domain graph convolution \cite{bruna2013spectral,defferrard2016convolutional,kipf2016semi} and spatial domain graph convolution \cite{hamilton2017inductive,velivckovic2017graph}. In the field of traffic prediction, spectral domain graph convolution is often used for spatial modeling. \cite{bruna2013spectral} proposed spectral domain graph convolution based on spectral graph theory for the first time. However, the eigenvalue decomposition of the Laplacian matrix has high computational complexity. \cite{defferrard2016convolutional} proposed ChebNet, which uses Chebyshev polynomials instead of convolution kernels, effectively reducing the huge computational effort of eigenvalue decomposition of Laplace matrices. \cite{kipf2016semi} further simplified ChebNet.

\section{Preliminaries}
In this section, we first present the associated definitions and formulation of the traffic flow prediction problem.  Then neural networks relevant to our work are introduced: graph convolution and self-attention mechanism. 

\subsection{Definition and Problem Statement}
{\noindent \bf Definition 1: Traffic network.}
As the prior knowledge of traffic flow prediction, the traffic road network can be represented as a graph $G=(V, E, A)$, where $V$ represents $N=|V|$ nodes in the traffic road network (e.g. observation points, road segments ), $E$ is the set of edges, and $A\in {{\mathbb{R}}^{N \times N}}$ denotes the adjacency matrix of correlations between nodes.

{\noindent \bf Definition 2: Traffic flow.}
The traffic flow of $N$ nodes at time $t$ can be expressed as $X^{(t)}=[X^t_1, X^t_2, \dots, X^t_N] \in {{\mathbb{R}}^{N\times C}}$, and $C$ is the traffic flow dimension of each node. $X^{(0:T)}=[X^{(0)}, X^{(1)}, \dots, X^{(T-1)}] \in {{\mathbb{R}}^{T \times N \times C}}$ represents the traffic flow tensor of $N$ nodes on the time slice $T$ in the traffic road network $G$.

The traffic forecasting task aims to combine prior knowledge and make full use of the complex spatial-temporal information in historical traffic data to achieve accurate forecasts of future traffic flows. The traffic forecasting problem can be expressed as learning the forecasting function $F$ from the past $T$ steps of traffic flow $X^{(t-T:t)}=[X^{(t-T)}, \dots, X^{(t-1)}]$ and the road network graph $G$ to forecast the traffic flow $X^{(t:{t+{T}^{'}})}=[X^{(t)}, \dots, X^{(t+{T}^{'}-1)}]$ at the next $T'$ steps:
\begin{equation}\label{eq1}
[X^{(t-T)}, \dots, X^{(t-1)}, G]\stackrel{F_{\Theta}}{\longrightarrow}[X^{(t)}, \dots, X^{(t+{T}^{'}-1)}],
\end{equation}
where $\Theta$ denotes all the learnable parameters in the prediction function F.

\subsection{ChebNet}
Our work uses ChebNet \cite{defferrard2016convolutional} for spatial modeling, which uses Chebyshev polynomials instead of convolution kernels. This can effectively reduce the computational cost of eigendecomposition of Laplacian matrices:
\begin{equation}\label{eq2}
Z = g_{\theta }\star_{G}x = \sum_{k=0}^{K-1}\theta_{k}T_{k}(\tilde{L})x,
\end{equation}
where $\theta_{k}$ is the learnable parameter and $K \geq 2$ is the number of convolution kernels. $\tilde{L} = \frac{2}{\lambda_{max}}L-I$ is the scaled Laplacian matrix, where $\lambda_{max}$ is the largest eigenvalue , $L=I-D^{-\frac{1}{2}}{A}D^{-\frac{1}{2}} \in \mathbb {R}^{N \times N}$ is the symmetric normalized graph Laplacian matrix and $I$ is the identity matrix. The Chebyshev polynomial is defined as $T_{0}(\tilde{L}) = I, T_{1}(\tilde{L}) = \tilde{L}$, and $T_{n+1}(\tilde{L}) = 2\tilde{L}T_{n}(\tilde{L}) - T_{n-1}(\tilde{L})$.

\subsection{Self-attention}
Self-attention is a variant of the attention mechanism that reduces reliance on external information and enhances the capture of relevance within the data. Specifically, query $Q=XW_{q} \in \mathbb {R}^{T \times d_{q}}$, key $K=XW_{k} \in \mathbb {R}^{T \times d_{k}}$, and value $V=XW_{v} \in \mathbb {R}^{T \times d_{v}}$ of the self-attendant are all obtained by linear transformations of the same matrix $X \in {{\mathbb{R}}^{T \times d}}$. where $d_{q}$, $d_{k}$, and $d_{v}$ are the dimensions of $Q$, $K$, and $V$, and $W_{q},W_{k},W_{v} \in \mathbb {R}^{d \times d}$ are the learnable parameters of the linear projection (our work preserves $d_q=d_k=d_v=d$). The formula can be expressed as follows:
\begin{equation}\label{eq3}
Att(Q, K, V) = softmax(\frac{({XW_{q}})({XW_{k}})^{\mathbf{T}}}{\sqrt{d}})({XW_{v}}).
\end{equation}

\section{Methodology}
\begin{figure*}[t]
\centering
\includegraphics[width=\textwidth]{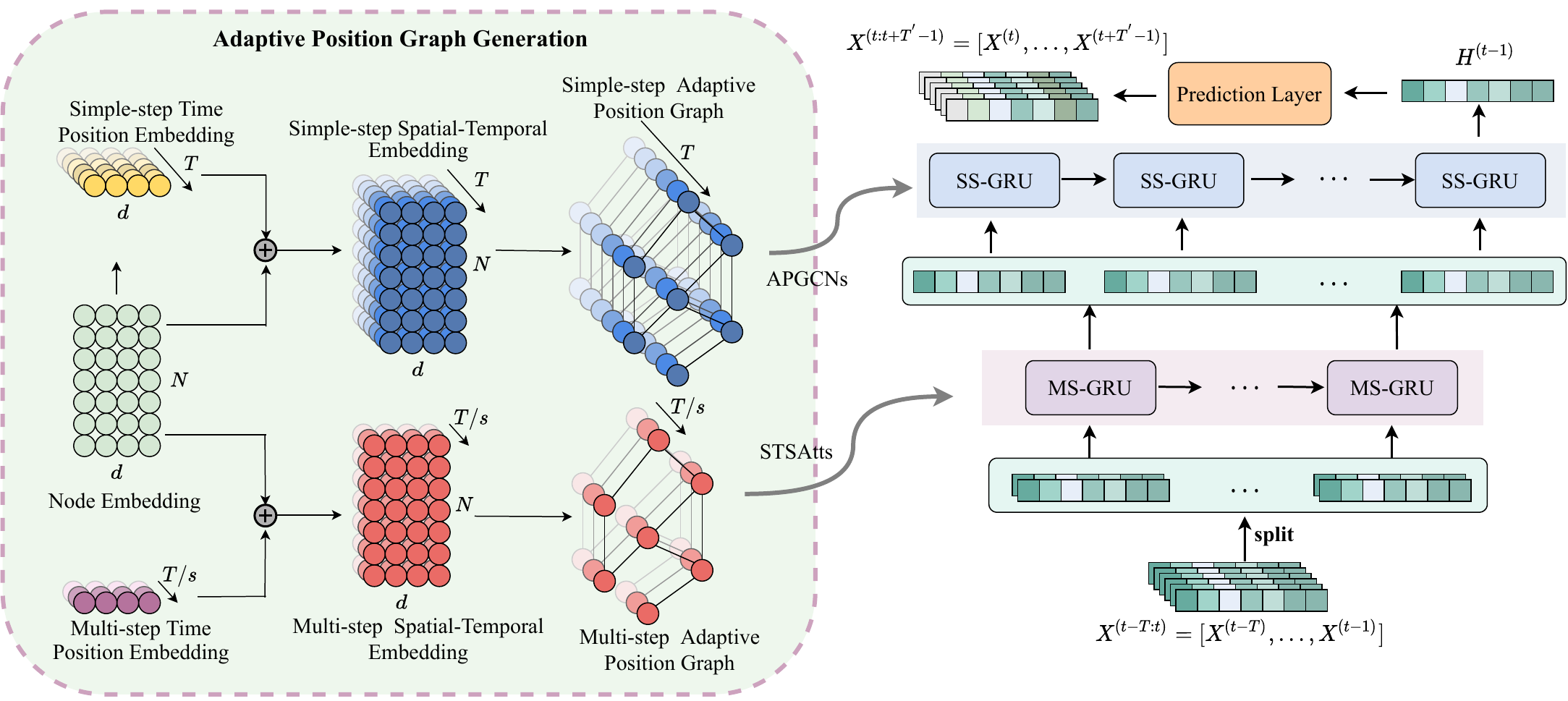} % Reduce the figure size so that it is slightly narrower than the column.
\caption{The framework of the MSSTRN model, with the adaptive position graph generation (APGG) module highlighted in the dashed box.}
\label{fig2}
\end{figure*}

In this section, we describe the detailed implementation of MSSTRN. The overall framework of MSSTRN is shown in Figure \ref{fig2}. It consists of three main modules: the adaptive position graph generation (APGG), the Multi-Step Gate Recurrent Unit (MS-GRU), and the Simple Step Gate Recurrent Unit (SS-GRU).

\subsection{Adaptive Position Graph Convolutions}
Recently adaptive adjacency matrices have been favored by spatial-temporal prediction researchers to capture more complex spatial dependencies by initializing a learnable adjacency matrix without prior knowledge \cite{bai2020adaptive,yu2022regularized}. However, they are premised on the fixed edge weights between nodes. To this end, we propose a adaptive position graph generation (APGG) module for generating each term of Chebyshev polynomials of ChebNet \cite{defferrard2016convolutional} to extract more meaningful and implicit spatial information on the traffic network.

\begin{figure}[!t]
\centering
\includegraphics[width=\columnwidth]{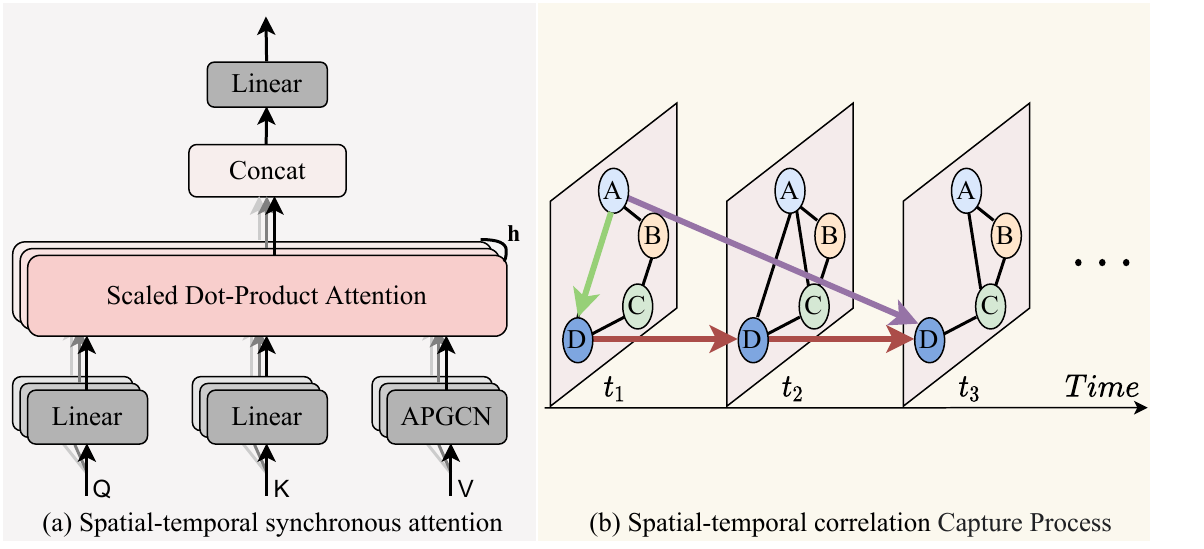} % Reduce the figure size so that it is slightly narrower than the column.
\caption{(a) Spatial-Temporal Synchronized Attention (STSAtt) mechanism. (b) Example of synchronous capture of spatial-temporal correlations.}
\label{fig3}
\end{figure}

{\noindent \bf Adaptive Position Graph Generation.}
\noindent Inspired by adaptive graph generation \cite{bai2020adaptive}, we can initialise a spatial-temporal embedding $E \in \mathbb {R}^{T \times N \times d}$ and then generate adaptive position graphs by permutation multiplication ($E \cdot E^{\mathbf{T}}$). However, the parametric number of $E$ is $O(TNd)$ and may lead to overfitting. Instead of learning $E$ directly, two different types of embeddings are learned using a parameter decomposition approach. We first randomly initialise a learnable node embedding $E_{\phi}\in \mathbb {R}^{N \times d_{\phi}}$ denoting the spatial information of each node, where $d_{\phi}$ denotes the size of the node embedding. Traffic flows of length $T$ can be decomposed into $T_c = T/s$ smaller windows along the time dimension by the window size $s$. We observe that time series with different window sizes have different spatial-temporal information. To this end, we initialize two types of time embeddings: the initial \emph{single-step} time position embedding $E_1\in \mathbb {R}^{T \times 1 \times d_{s}}$ and the \emph{multi-step} time position embedding $E_2\in \mathbb {R}^{T_{c} \times 1 \times d_{m}}$ that contains relative temporal order information. To facilitate model manipulation, we set $d_{\phi}=d_{s}=d_{m}=d$ in our work. At this time, the number of parameters is reduced to $O(Nd+Td+T_{c}d)<O(TNd+T_{c}Nd)$.
% The single-step node embedding $E_{s}=LayerNorm(E_{\phi}+E_1) \in \mathbb {R}^{T \times N \times d_{s}}$ and the multi-step node embedding $E_{m}=LayerNorm(E_{\phi}+E_2) \in \mathbb {R}^{T_{c} \times N \times d_{m}}$ are two node embeddings with information about the dynamic of the node itself at different time windows. The layer normalization $LayerNorm(\cdot)$ \cite{ba2016layer} ensures the stability of $E_s$ and $E_m$ during the training process. 
The scaled Laplacian matrix for time step $i$ ($s=1$):
% \begin{equation}\label{eq4}
% S_{1}(\hat{L})[i] = softmax(E_{s}[i]\cdot E_{s}[i]^{\mathbf{T}}),
% \end{equation}
\begin{equation}
    \begin{split}\label{eq4}
    S_{1}(\hat{L})[i] &= softmax(E_{s}[i]\cdot E_{s}[i]^{\mathbf{T}}), \\
    \text{where}\ E_{s}[i]&=LayerNorm(E_{\phi}+E_1[i]),
    \end{split}
\end{equation}
and the scaled Laplacian matrix for time window $j$ ($s>1$):
% \begin{equation}\label{eq5}
% M_{1}(\hat{L})[j] =  softmax(E_{m}[j]\cdot E_{m}[j]^{\mathbf{T}}).
% \end{equation}
\begin{equation}
    \begin{split}\label{eq5}
    M_{1}(\hat{L})[j] =  softmax(E_{m}[j]\cdot E_{m}[j]^{\mathbf{T}}), \\
    \text{where}\ E_{m}[j]=LayerNorm(E_{\phi}+E_2[j]),
    \end{split}
\end{equation}
where the single-step spatial-temporal embedding $E_{s} \in \mathbb {R}^{T \times N \times d_{s}}$ and the multi-step spatial-temporal embedding $E_{m} \in \mathbb {R}^{T_{c} \times N \times d_{m}}$ are two node embeddings with information about the nodes in different time windows at different time positions. The layer normalization $LayerNorm(\cdot)$ \cite{ba2016layer} ensures the stability of $E_s$ and $E_m$ during the training process.

To explore the hidden spatial correlations between node fields of different depths, we generate ${S}_{k}(\hat{L})$ and ${M}_{k}(\hat{L})$ of different depths $k$ based on Chebyshev polynomials and concatenate them as tensors $\tilde{T}_{s} = [I, {S}_{1}(\hat{L}), \dots, {S}_{K-1}(\hat{L})]^{\mathbf{T}}\in \mathbb {R}^{K \times T \times N \times N}$ and $\tilde{T}_{m} = [I, {M}_{1}(\hat{L}), \dots, {M}_{K-1}(\hat{L})]^{\mathbf{T}}\in \mathbb {R}^{K \times T_{c} \times N \times N}$.

{\noindent \bf Adaptive Position Graph Convolution Networks.}
\noindent We bring $\tilde{T}_{s}$ and $\tilde{T}_{m}$ into Eq. \eqref{eq2} and choose ChebNet with independent node parameters \cite{bai2020adaptive} for convolution operation. Therefore, the formula for the convolution operation on the graph signal $X_i \in \mathbb {R}^{N \times d}$ at time $i$:
\begin{equation}\label{eq6}
G(X^{(i)}) = {\tilde{T}_{s}[:,i]}X^{(i)}{E_{s}[i]}W_i+{E_{s}[i]}b_i,
\end{equation}
and the graph convolution formula for the graph signal $X_{j:j+s} \in \mathbb {R}^{s \times N \times d}$ with window size $s$:
\begin{equation}\label{eq7}
G(X^{(j:j+s)}) = {\tilde{T}_{m}[:,j]}{X^{(j:j+s)}}{E_{m}[j]}W_j+{E_{m}[j]}b_j,
\end{equation}
where $W_i, W_j \in \mathbb {R}^{d \times K \times d_{in} \times d_{out}}$ and $b_i, b_j \in \mathbb {R}^{d \times d_{out}}$ are the learnable parameters. Our working default setting $d_{in}=d_{out}=d$.

\begin{figure*}[t]
\centering
\includegraphics[width=\textwidth]{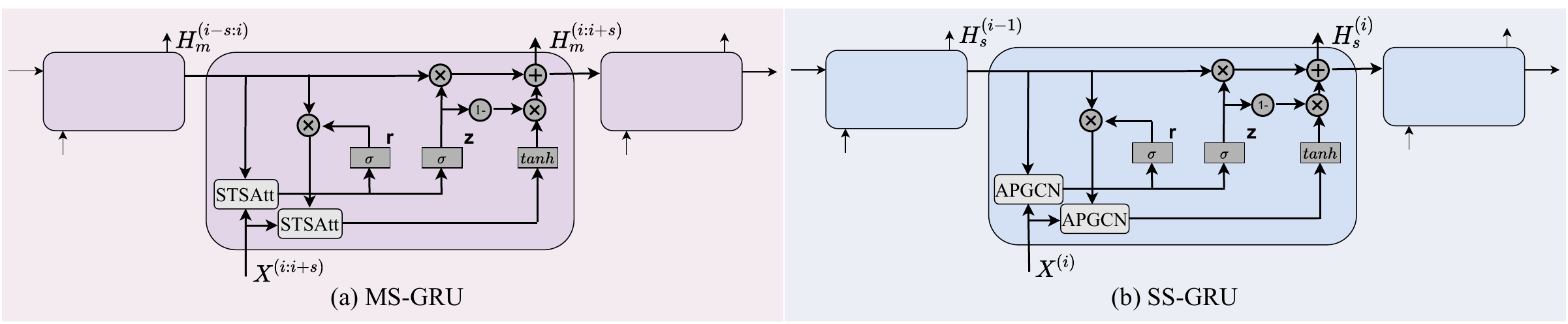} % Reduce the figure size so that it is slightly narrower than the column.
\caption{Spatial-temporal Recurrent Neural Networks. (a) is the Multi-Step Gate Recurrent Unit (MS-GRU), which uses Spatial-Temporal Synchronous Attention (STSAtt) to replace the fully connected operation of the GRU. (b) is the Single-Step Gate Recurrent Unit (SS-GRU), which uses adaptive position graph convolution (APGCN) to replace the fully connected operation of the GRU.}
\label{gru}
\end{figure*}

\subsection{Spatial-Temporal Synchronous Attention}
As we described in Section I, information is often simultaneously conveyed in both temporal and spatial dimensions, necessitating simultaneous spatial-temporal modeling. Current approaches using spatial-temporal graphs to capture spatial-temporal dependencies have limited perceptual field and are difficult to capture long-term correlations. We propose a spatial-temporal attention mechanism that can synchronously capture complex spatial-temporal correlations in road networks: the Spatial-Temporal Synchronous Attention (STSAtt) Mechanism (see Figure \ref{fig3}(a)),  a combination of self-attention and adaptive position graph convolution.

Unlike Equation \ref{eq3}, the value $V=G(X)$ of STSAtt is obtained by a adaptive position graph convolution operation. Specifically, the APGCN first makes the value $V$ spatially informative and then uses the temporal attention scores of $Q$ and $K$ to achieve synchronous capture of spatial-temporal correlations. For example in Figure \ref{fig3}(b), the impact from node $A$ at time point $t_1$ to node $D$ at time point $t_3$ (in purple arrow) is first captured by the spatial relationship from $A$ to $D$ at the same time point $t_1$ (in green arrow) and then by the temporal relationship from $D$ at time point $t_3$ (in red arrow). To learn the dependencies of the different patterns, we use a multi-headed attention mechanism, and the formula can be written as:
\begin{equation}
    \begin{split}\label{eq8}
    A(X) &= Concat(head_{1}, \dots, head_{h})W_o, \\
    \text{where}\ head_{j} &= Att(XW^j_q, XW^j_k, G(X)),
    \end{split}
\end{equation}
where $X \in \mathbb {R}^{N \times T \times d_{in}}$ is the input, $h$ is the number of heads, and the learnable parameters $W_o \in \mathbb {R}^{d_{out} \times d_{out}}$, $W^j_{q},W^j_{k} \in \mathbb {R}^{d_{in} \times d_{out}}$.

\subsection{Spatial-temporal Recurrent Neural Networks}
As shown in Fig. \ref{fig2}, we first initialize the node embedding $E_{\phi}$ and the time position embeddings $E_1$ and $E_2$ to generate the learnable scaled Laplace tensors $\tilde{T}_{m}$ and $\tilde{T}_{s}$. To capture the unique interdependencies between time slices, we split the time series $X^{(t-T:t)}$ into multiple sub-windows ($s>1$) and input them simultaneously with $\tilde{T}_{m}$ as inputs to Multi-Step Gate Recurrent Unit (MS-GRU, see Figure \ref{gru}(a)) formed by replacing MLP layers in the GRU with STSAtt. The specific operation can be formulated as follows:
\begin{equation}
    \begin{split}\label{eq9}
    z^{(i:i+s)} &= \sigma(A([X^{(i:i+s)},h^{(i-s:i)}])), \\
    r^{(i:i+s)} &= \sigma(A([X^{(i:i+s)},h^{(i-s:i)}])), \\
    \tilde{h}^{(i:i+s)} &= tanh(A([X^{(i:i+s)},r^{(i:i+s)} \odot h^{(i-s:i)}])), \\
    h^{(i:i+s)} &= z^{(i:i+s)}\odot h^{(i-s:t)} + (1-z^{(i:i+s)})\odot \tilde{h}^{(i:i+s)},
    \end{split}
\end{equation}
where $\sigma$ and $tanh$ are two activation functions, i.e., the Sigmoid function and the Tanh function. The $[X^{(i:i+s)},h^{(i-s:i)}]$ and $h^{(i:i+s)}$ are the input and output at time window $[i:i+s]$, respectively. 

To further discover the spatial-temporal correlation between the time series ($s = 1$), we take the output ${H_{m}}^{(t-T:t)}$ of MS-GRU as input and input it simultaneously with $\tilde{T}_{s}$ to Single-Step Gate Recurrent Unit (SS-GRU, see Fig. \ref{gru}(b)) formed by replacing MLP layers in the GRU with APGCN. The operation can be formulated as follows:
\begin{equation}
    \begin{split}\label{eq10}
    z^{(i)} &= \sigma(G([{H_{m}}^{(i)},h^{(i-1)}])), \\
    r^{(i)} &= \sigma(G([{H_{m}}^{(i)},h^{(i-1})])), \\
    \tilde{h}^{(i)} &= tanh(G([{H_{m}}^{(i)},r^{(i)} \odot h^{(i-1)}])), \\
    h^{(i)} &= z^{(i)}\odot h^{(i-1)} + (1-z^{(i)})\odot \tilde{h}^{(i)},
    \end{split}
\end{equation}
where $[{H_{m}}^{i},h^{i-1}]$ and $h^{i}$ are the input and output at time step $i$, respectively. The output of the SS-GRU is $H_s^{(t-T:t)}$. Considering the cumulative error and efficiency issues of RNNs, we abandoned the autoregressive approach and chose to perform multi-step prediction directly. We use 2D convolution to transform the output $H_s^{(t-1)}$ of the final time step of the SS-GRU to the prediction dimension.
\begin{equation}\label{eq11}
X^{(t:{t+{T}^{'}})} = Conv2d(LayerNorm({H_s}^{(t-1)})),
\end{equation}
where $X^{(t:{t+{T}^{'}})} \in \mathbb {R}^{{T}^{'} \times N \times 1}$ is the prediction result and $Conv2d$ is a $1 \times {T}^{'}$ 2D convolution operation. It is worth noting that one may stack MS-GRU and SS-GRU in different ways. In Section V.H, we perform extensive experiments on this and summarize several interesting observations on the empirical performance.

\subsection{Loss Function}
We choose the $L1$ loss to train the prediction function $F_{\Theta}$ and minimize the training error by back propagation:
\begin{equation}\label{eq14}
Loss = \frac{1}{{T}^{'}} ({\hat{X}}^{(t:{t+{T}^{'}})}-X^{(t:{t+{T}^{'}})}),
\end{equation}
where ${\hat{X}}^{(t:{t+{T}^{'}})}$ is the ground-truth traffic data.

\section{Experimental Results}
In this section, we demonstrate the validity of MSSTRN through a series of experiments. We first describe the datasets and the experimental setup, and then analyse the predicted results obtained. Finally, we provide a comprehensive discussion of the ablation study and hyperparameter tuning.

\subsection{Datasets}
We evaluate the performance of our proposed model on four traffic datasets from the Caltrans Performance Measure System (PeMS) \cite{chen2001freeway}: PEMSD3, PEMSD4, PEMSD7 and PEMSD8 \cite{fang2021spatial}. Data preprocessing aggregates traffic data into 5-minute intervals, thus $288$ data points per day. In our experiments, we only use traffic flow data, ignoring traffic speed and traffic volume data. In addition, we use the Z-score normalization method to normalize all input data to enhance the stability of the training process. Detailed statistics for the 4 real traffic datasets are summarised in Table \ref{table1}.
\begin{table}[!t]
\centering
\caption{Statistics of The Tested Datasets}
\label{table1}
\begin{tabular}{c c c c c c}
    \toprule[0.8pt]
    % \hline
    Datasets & Nodes & Samples & Unit & Time Span \\
    \hline
    PEMSD3 & $358$ & $26,208$ & $5$ mins & $3$ months \\
    PEMSD4 & $307$ & $16,992$ & $5$ mins & $2$ months \\
    PEMSD7 & $883$ & $28,224$ & $5$ mins & $4$ months \\
    PEMSD8 & $170$ & $17,856$ & $5$ mins & $2$ months \\
    \bottomrule[0.8pt]
    % \hline
\end{tabular}
\end{table}

\begin{table*}[!t]
	\renewcommand{\arraystretch}{1.2}
	\centering
        \caption{Performance Comparison of Different Models on The Tested Datasets. Underlined Results Are The Current State of The Art Among The Existing Methods. Our Model Outperform All The Baseline Methods, as Shown in Bold Font.}
	\label{table2}
	\resizebox{\linewidth}{!}{
	\begin{tabular}{c|c c c|c c c |c c c|c c c}
		\toprule[0.8pt]
		\multirow{2}{*}{Model} & \multicolumn{3}{c|}{PEMSD3} & \multicolumn{3}{c|}{PEMSD4} & \multicolumn{3}{c|}{PEMSD7} & \multicolumn{3}{c}{PEMSD8} \\
		\cline{2-13}
		{}& MAE & RMSE & MAPE & MAE & RMSE & MAPE & MAE & RMSE & MAPE & MAE & RMSE & MAPE \\
		\midrule
		HA & 31.58 & 52.39 & 33.78\% & 38.03 & 59.24 & 27.88\% & 45.12 & 65.64 & 24.51\% & 34.86 & 59.24 & 27.88\% \\
		ARIMA & 35.41 & 47.59 & 33.78\% & 33.73 & 48.80 & 24.18\% & 38.17 & 59.27 & 19.46\% & 31.09 & 44.32 & 22.73\% \\
		VAR & 23.65 & 38.26 & 24.51\% & 24.54 & 38.61 & 17.24\% & 50.22 & 75.63 & 32.22\% & 19.19 & 29.81 & 13.10\% \\
		SVR & 20.73 & 34.97 & 20.63\% & 27.23 & 41.82 & 18.95\% & 32.49 & 44.54 & 19.20\% & 22.00 & 33.85 & 14.23\% \\ \hline
		FC-LSTM & 21.33 & 35.11 & 23.33\% & 26.77 & 40.65 & 18.23\% & 29.98 & 45.94 & 13.20\% & 23.09 & 35.17 & 14.99\%\\
		DCRNN(2018) & 17.99 & 30.31 & 18.34\% & 21.22 & 33.44 & 14.17\% & 25.22 & 38.61 & 11.82\% & 16.82 & 26.36 & 10.92\% \\
		AGCRN(2020) &15.98 & 28.25 & 15.23\% & 19.83 & 32.26 & 12.97\% & 22.37 & 36.55 & 9.12\% & 15.95 & 25.22 & 10.09\% \\ 
		Z-GCNETs(2021) & 16.64 & 28.15 & 16.39\% & 19.50 & 31.61 & 12.78\% & 21.77 & 35.17 & 9.25\% & 15.76 & 25.11 & 10.01\% \\ 
		RGSL(2022) & 15.85 & 28.51 & \underline{14.68\%} & 19.19 & 31.14 & 12.69\% & 20.58 & 33.88 & \underline{8.69\%} & 15.49 & 24.80 & 9.96\% \\
		GMSDR(2022) & 15.78 & 26.82 & 15.33\% & 20.37 & 32.52 & 13.71\% & 21.89 & 35.46 & 9.42\% & 16.36 & 25.58 & 10.28\% \\ \hline
		STGCN(2018) & 17.55 & 30.42 & 17.34\% & 21.16 & 34.89 & 13.83\% & 25.33 & 39.34 & 11.21\% & 17.50 & 27.09 & 11.29\% \\ 
		Graph WaveNet(2019) & 19.12 & 32.77 & 18.89\% & 24.89 & 39.66 & 17.29\% & 26.39 & 41.50 & 11.97\% & 18.28 & 30.05 & 12.15\% \\
		LSGCN(2020) & 17.94 & 29.85 & 16.98\% & 21.53 & 33.86 & 13.18\% & 27.31 & 41.46 & 11.98\% & 17.73 & 26.76 & 11.20\% \\
		STSGCN(2020) & 17.48 & 29.21 & 16.78\% & 21.19 & 33.65 & 13.90\% & 24.26 & 39.03 & 10.21\% & 17.13 & 26.80 & 10.96\% \\
		STFGNN(2021) & 16.77 & 28.34 & 16.30\% & 20.48 & 32.51 & 16.77\% & 23.46 & 36.60 & 9.21\% & 16.94 & 26.25 & 10.60\% \\ \hline
		ASTGCN(r)(2019) & 17.34 & 29.56 & 17.21\% & 22.93 & 35.22 & 16.56\% & 24.01 & 37.87 & 10.73\% & 18.25 & 28.06 & 11.64\% \\
		DSTAGNN(2022) &  15.57 & 27.21 &  \underline{14.68\%} & 19.30 & 31.46 &  12.70\% & 21.42 & 34.51 & 9.01\% & 15.67 &  24.77 & 9.94\% \\ 
		ST-WA(2022) & \underline{15.17} & \underline{26.63} & 15.83\% & \underline{19.06} & \underline{31.02} & \underline{12.52\%} & 20.74 & 34.05 & 8.77\% & \underline{15.41} & \underline{24.62} & 9.94\%\\ \hline
		STGODE(2021) & 16.50 & 27.84 & 16.69\% & 20.84 & 32.82 & 13.77\% & 22.59 & 37.54 & 10.14\% & 16.81 & 25.97 & 10.62\% \\
		STG-NCDE(2022) &  15.57 &  27.09 & 15.06\% &  19.21 &  31.09 & 12.76\% &  \underline{20.53} &  \underline{33.84} &  8.80\% &  15.45 & 24.81 &  \underline{9.92\%}\\
		\midrule
		\bf{MSSTRN} (Ours) & \bf{14.83} & \bf{26.25} & \bf{14.04}\% & \bf{18.77} & \bf{30.91} & \bf{12.14}\% & \bf{20.02} & \bf{33.34} & \bf{8.39}\% & \bf{14.92} & \bf{24.17} & \bf{9.59}\% \\
		\bf{Improvements} & \emph{+2.24\%} & \emph{+1.43\%} & \emph{+4.36\%} & \emph{+1.52\%} & \emph{+0.45\%} & \emph{+3.04\%} & \emph{+2.48\%} & \emph{+1.48\%} & \emph{+3.45\%} & \emph{+3.18\%} & \emph{+1.83\%} & \emph{+3.33\%} \\
		\bottomrule[0.8pt]
	\end{tabular}
	}
\end{table*}

\begin{table}[!t]
	\renewcommand{\arraystretch}{1.2}
	\centering
        \caption{Computation Time and Memory Cost of MSSTRN and Several Recent Competitive Methods on PEMSD4.}
	\label{table3}
	\resizebox{.70\columnwidth}{!}{
	\begin{tabular}{c|c c c}
		\toprule[0.8pt]
        Model &  Train. & Infer. & Mem. \\
		\midrule
		STGODE(2021) & 111.77 & 12.19 & 8773 \\
		Z-GCNETs(2021) & 63.34 & 7.40 & 8597 \\
		DSTAGNN(2022) &  242.57 & 14.64 & 10347 \\
		STG-NCDE(2022) & 1318.35 & 93.77 & 6091 \\
		RGSL(2022) & 99.36 & 20.51 & 6843 \\
		GMSDR(2022) & 125.19 & 19.20 &  5751 \\
		ST-WA(2022) & 84.43 & 4.92 & 4305 \\
		\midrule
		\bf{MSSTRN} & 40.02 & 4.71 & 4752 \\
		\bottomrule[0.8pt]
	\end{tabular}
	}
\end{table}

\subsection{Baseline Methods}
In order to fully evaluate the performance of our models, we selected 20 baseline methods and grouped them into five categories:
\begin{itemize}
    \item Traditional time series forecasting methods: (1) Historical Average (\textbf{HA}), which uses the average value of historical traffic flows to predict future traffic flows; (2) \textbf{ARIMA} \cite{williams2003modeling}, which is a widely used model for time series forecasting; (3) \textbf{VAR} \cite{zivot2006vector}, a statistical model that captures the relationship between multiple variables over time and (4) \textbf{SVR} \cite{drucker1996support}, which uses linear support vector machines for regression tasks. 
    \item RNN-based models: (5) \textbf{FC-LSTM} \cite{sutskever2014sequence}, LSTM network with fully connected hidden units; (6) Diffusion Convolutional Recurrent Neural Network (\textbf{DCRNN}) \cite{li2018dcrnn_traffic}, which captures spatial and temporal dependencies using diffuse graph convolution and encoder-decoder network architecture, respectively; (7) Adaptive Graph Convolutional Recurrent Network (\textbf{AGCRN}) \cite{bai2020adaptive}, which augments traditional graph convolution with adaptive graph generation and node adaptive parameter learning, and is integrated into a recurrent neural network to capture more complex spatial-temporal correlations; (8) Time Zigzags at Graph Convolutional Networks (\textbf{Z-GCNETs}) \cite{chen2021z}, which introduces the concept of Zigzag persistence to time-aware graph convolutional networks; (9) Regularized Graph Structure Learning model (\textbf{RGSL}) \cite{yu2022regularized}, which first introduces the Regularized Graph Generation module to learn implicit graphs, and then the Laplacian Matrix Mixed-up module to combine explicit and implicit structures and (10) Graph-based Multi-Step Dependency Relation (\textbf{GMSDR}) \cite{liu2022msdr}, integrating graph neural networks with Multi-Step Dependency Relation, a new variant of recurrent neural networks, for spatial-temporal prediction.
    \item CNN-based methods: (11) Spatial-Temporal Graph Convolutional Network (\textbf{STGCN}) \cite{yu2018spatio}, which combines graph convolution and 1D convolution to capture spatial-temporal correlations; (12) \textbf{Graph WaveNet} \cite{wu2019graph}, which introduces an adaptive adjacency matrix and combines diffuse graph convolution with 1D convolution; (13) Long Short-term Graph Convolutional Networks (\textbf{LSGCN}) \cite{huang2020lsgcn}, which proposes a new graph attention network and integrates it with graph convolution into a spatial gated block; (14) Spatial-Temporal Synchronous Graph Convolutional Networks (\textbf{STSGCN}) \cite{song2020spatial}, which enables the model to efficiently extract localized spatial-temporal correlations through a well-designed local spatial-temporal subgraph module and (15) Spatial-Temporal Fusion Graph Neural Networks (\textbf{STFGNN}) \cite{li2021spatial}, which designs a new spatial-temporal fusion graph module and assembles it in parallel with 1D convolution module.
    \item Attention-based models: (16) Attention Based Spatial-Temporal Graph Convolutional Networks (\textbf{ASTGCN(r)}) \cite{guo2019attention},  which fuses spatial attention and temporal attention mechanisms with spatial-temporal convolution to capture dynamic spatial-temporal features; (17) Dynamic Spatial-Temporal Aware Graph Neural Network (\textbf{DSTAGNN}) \cite{lan2022dstagnn}, which proposes a new dynamic spatial-temporal awareness graph to replace the predefined static graph used by traditional graph convolution; and (18) (\textbf{ST-WA}) \cite{cirstea2022towards}, which consists of multiple layers of Spatial-Temporal Aware Window Attention.
    \item Differential equation-based methods: (19) Spatial-Temporal Graph Ordinary Differential Equation Networks (\textbf{STGODE}) \cite{fang2021spatial}, which captures spatial-temporal dynamics through a tensor-based ordinary differential equation (ODE) and (20) Spatial-Temporal Graph Neural Controlled Differential Equation (\textbf{STG-NCDE}) \cite{choi2022graph}, which designs two NCDEs for temporal processing and spatial processing and integrates them into a single framework. 
\end{itemize}

\subsection{ Experimental Settings}
Keeping consistent with the baseline methods, the tested datasets were split into training, validation and test data in a $6:2:2$ ratio. Our model uses traffic data from the last $12$ continuous time steps ($1$ hour) to predict the traffic flow for the next $12$ continuous time steps. 

MSSTRN was implemented using the PyTorch framework, and the series of experiments was performed on an NVIDIA GTX 1080 TI GPU with 11GB of memory. The following hyperparameters were configured based on the model's predictive performance on the validation set: we used the Adam optimiser \cite{kingma2014adam} with a learning rate of $0.003$ and a batch size of $64$ for all datasets. we set epochs to $500$ and adopted an early stopping strategy with a patience number of $30$. The number of convolution kernels of APGCN $K = 2$ and the number of heads of STSAtt $h = 4$. The weight decay coefficients were chosen from $w \in {\{0, 0.0001, \cdots, 0.001\}}$, node embedding size $d \in {\{1, 2, \cdots, 10\}}$, and the size of sub-windows $s \in {\{2, 3, 4, 6\}}$.

Three common prediction metrics, Mean Absolute Error (MAE), Root Mean Square Error (RMSE), and Mean Absolute Percentage Error (MAPE), are used to measure the traffic forecasting performance \cite{chen2021z}. Their formal definitions are as follows:
\begin{equation}
\begin{aligned}
&\operatorname{MAE}(\hat{Y}, Y)=\frac{1}{T} \sum_{i=1}^{T}\left|\hat{y}_{i}-y_{i}\right| \\
&\operatorname{RMSE}(\hat{Y}, Y)=\sqrt{\frac{1}{T} \sum_{i=1}^{T}\left(\hat{y}_{i}-y_{i}\right)^{2}} \\
&\operatorname{MAPE}(\hat{Y}, Y)=\frac{100\%}{T} \sum_{i=1}^{T}\left|\frac{\hat{y}_{i}-y_{i}}{\hat{y}_{i}}\right|
\end{aligned}
\end{equation}
where $\hat{Y}=\hat{y}_{1}, \hat{y}_{2}, \dots, \hat{y}_{T}$ is the real traffic data, $Y={y}_{1}, {y}_{2}, \dots, {y}_{T}$ is the predicted data, and $T$ is the predicted time step. In our experiments, $T=12$.

\begin{figure*}[!t]
\centering
\includegraphics[width=0.9\linewidth]{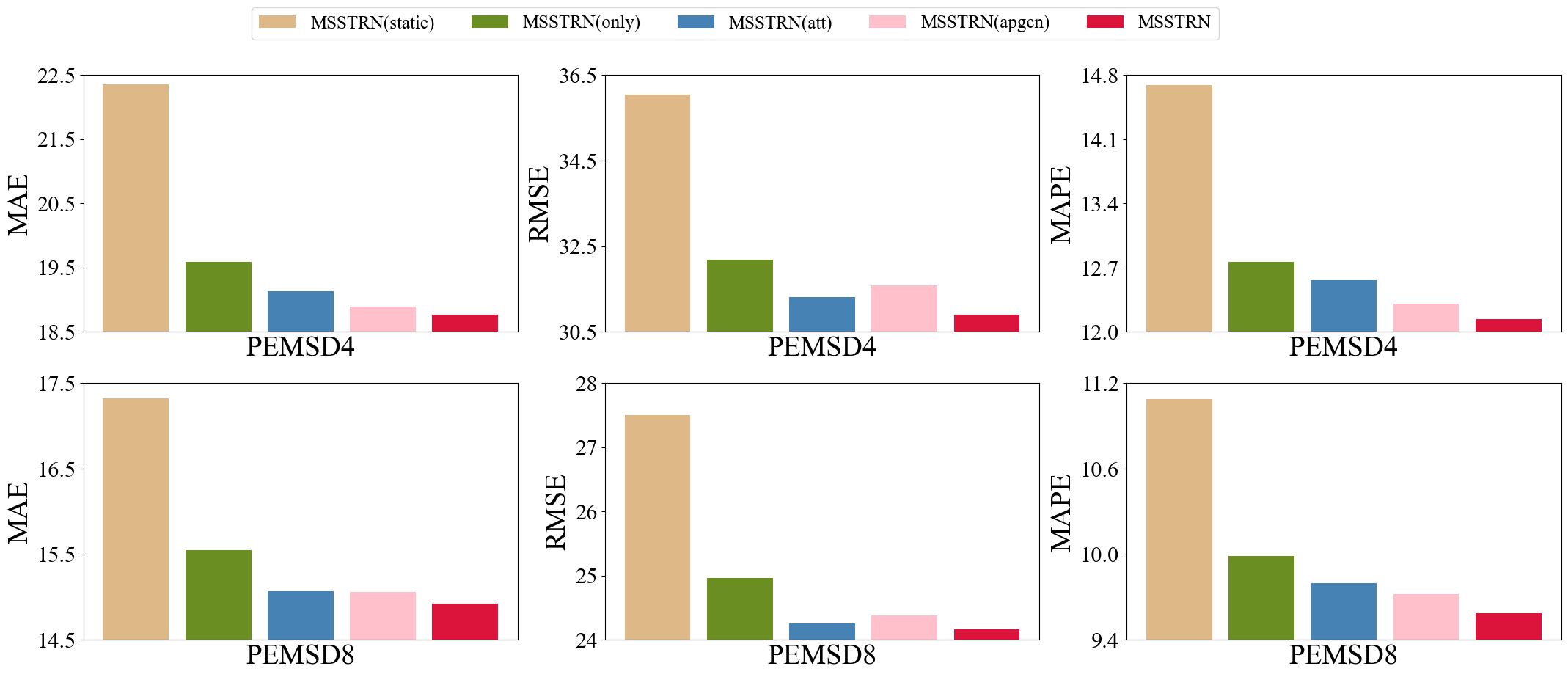}
\caption{Ablation experiment on PEMSD4 and PEMSD8.}
\label{fig4}
\end{figure*}

\subsection{Experimental Results}
Table \ref{table2} shows the predictive performance of our model and the twenty baseline methods on the four tested datasets. It is worth noting that our model consistently outperforms all the baseline methods on every dataset. In particular, on the PEMSD3 and PEMSD8 datasets, our model significantly improves the predictive performance of the current state-of-the-art methods by a non-trivial margin. The MAE, RMSE, and MAPE values outperform the state-of-the-art methods by $2.24\%$, $1.43\%$, and $4.36\%$ respectively on the PEMSD3 dataset. Similarly, on the PEMSD8 dataset, these three values are improved by $3.18\%$, $1.83\%$, and $3.33\%$ respectively. It is evident that the proposed framework can effectively model complex spatial-temporal correlations in traffic road networks with superior performance compared to all the baselines. 

Traditional statistical methods (including HA, ARIMA, VR, and SVR) have limited ability to deal with nonlinear data, so the prediction performance is much worse than that of deep learning methods. FC-LSTM uses only LSTM for temporal modeling and has the worst prediction performance among deep learning methods. Our model outperforms compared to RNN-based methods such as DCRNN, AGCRN, Z-GCNETs, RGSL and GMSDR, which use a sequence-to-sequence approach for multi-step prediction, where the accumulation of errors affects the model prediction performance. Although the autoregressive approach is abandoned and direct multi-step prediction is chosen, the generation of stationary graphs limits the performance of the model. In our empirical study, CNN-based models such as STGCN, Graph WaveNet, STSGCN, and STFGCN have worse or comparable performance compared to RNN-based methods. Although ST-WA has good prediction results, it uses only the attention module for spatial-temporal modeling and ignores the advantages of graph convolution for spatial modeling, which may limit its prediction performance to some extent. The good results from STG-NCDE show that differential equation-based spatial-temporal modeling has great research potential.

{\noindent \bf Run cost:} Table \ref{table3} lists the training time (s/epoch), inference time (s/epoch) and memory cost (MB) of our model and several recent and best-performing baselines on the PEMSD4 dataset. Except for ST-WA, our MSSTRN outperforms other state-of-the-art recent works in all three cost metrics. Although MSSTRN consumes more memory compared to ST-WA, the training time is faster.

% \begin{figure*}[!t]
% \centering
% \subfloat[]{\includegraphics[width=2.3in]{figures/PEMSD4_MAE.png}%
% \label{fig_a}}
% \hfil
% \subfloat[]{\includegraphics[width=2.3in]{figures/PEMSD4_RMSE.png}%
% \label{fig_b}}
% \hfil
% \subfloat[]{\includegraphics[width=2.3in]{figures/PEMSD4_MAPE.png}%
% \label{fig_c}}

% \subfloat[]{\includegraphics[width=2.3in]{figures/PEMSD8_MAE.png}%
% \label{fig_d}}
% \hfil
% \subfloat[]{\includegraphics[width=2.3in]{figures/PEMSD8_RMSE.png}%
% \label{fig_e}}
% \hfil
% \subfloat[]{\includegraphics[width=2.3in]{figures/PEMSD8_MAPE.png}%
% \label{fig_f}}
% \caption{Ablation experiments. Comparison of the prediction performance of each horizon on PEMSD4 and PEMSD8.}
% \label{fig_sim}
% \end{figure*}

\begin{figure*}[!t]
\centering
\subfloat[]{\includegraphics[width=3.5in]{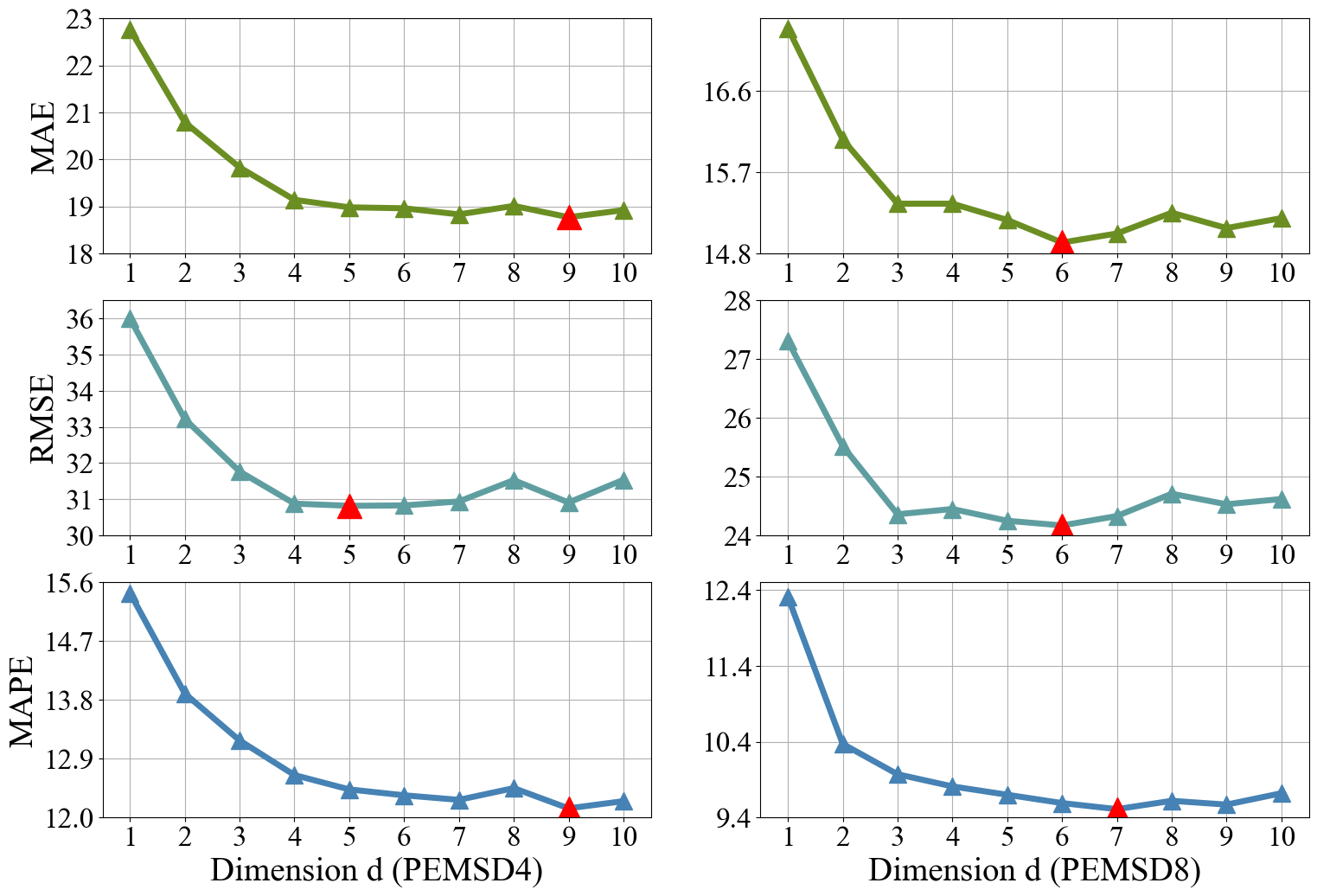}%
\label{fig_parar1}}
\hfil
\subfloat[]{\includegraphics[width=3.5in]{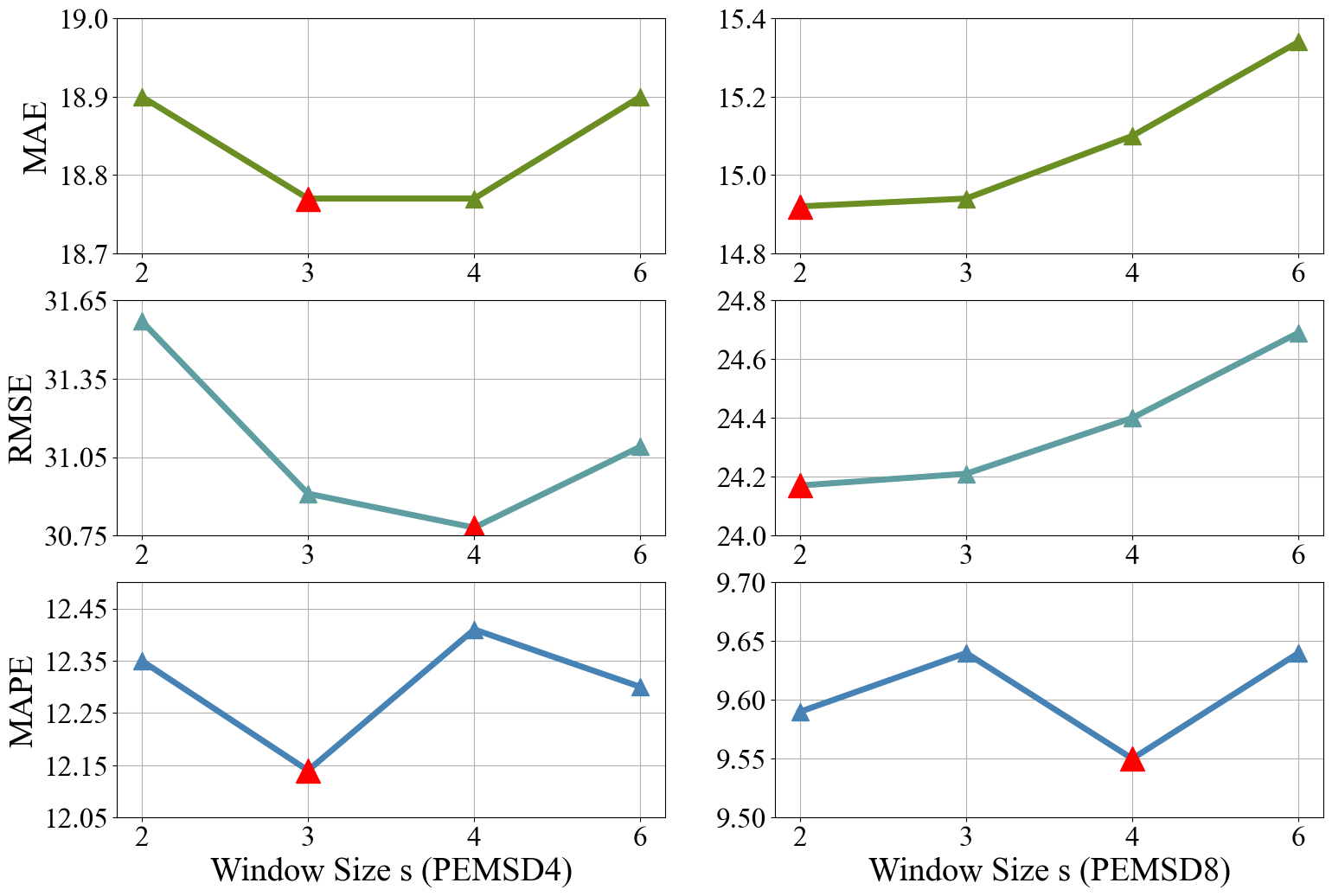}%
\label{fig_parar2}}
\caption{Hyperparameter Study on PEMSD4 and PEMSD8. (a) Effect of the dimension of node embedding $E_{\phi}$. (b) Effect of Window Size $s$.}
\label{fig_parar}
\end{figure*}

\subsection{Ablation Study}
To show the validity of the different components in the MSSTRN, a comprehensive ablation study was carried out on the PEMSD4 and PEMSD8 datasets, including: (1) {MSSTRN(static)}, using a predefined adjacency matrix; (2) MSSTRN(only), initializing the learnable adjacency matrix using only the node embedding $E_{\phi}$; (3) MSSTRN(att), replacing STSAtt in MSSTRN using a multi-headed attention mechanism; (4) MSSTRN(apgcn), replacing STSAtt in MSSTRN using adaptive position graph convolution.

All variant models are set to the same parameters as MSSTRN. Fig. \ref{fig4} shows the results of ablation experiments, confirming the necessity of each component in our model. Firstly, MSSTRN(only) improves the accuracy of {MSSTRN(static)}. It is clear that the adaptive adjacency matrix does outperform static graphs. However, the ability of APGG in MSSTRN to capture spatial dependencies is far better than using static graphs or adaptive adjacency matrices. Furthermore, by comparing MSSTRN(att), MSSTRN(apgcn), and MSSTRN, we can see the ability of STSAtt to simultaneously capture spatial-temporal dependencies. 

% As shown in Fig.\ref{fig_sim}, we plot the detailed values of the three metrics for the different horizons for the ablation model. Compared to the four after-ablation variants, MSSTRN consistently has the lowest values for the three metrics for all horizons on PEMSD4 and PEMS08 datasets.

\begin{table}[!t]
	\renewcommand{\arraystretch}{1.2}
	\centering
 	\caption{Effect of Convolution Kernel Number $K$.}
	\label{table6}
	\resizebox{.95\columnwidth}{!}{
	\begin{tabular}{c|c|c c c c c c}
		\toprule[0.8pt]
		Dataset& {$K$}& MAE & RMSE & MAPE & Train. & Infer. & Mem. \\
		\midrule
		\multirow{3}{*}{PEMSD4} & \bf{2} & \bf{18.77} & 30.91 & \bf{12.14}\% & \bf{40.02} & \bf{4.71} & \bf{4752} \\
		{} & 3 & \bf{18.77} & \bf{30.82} & 12.21\% & 50.38 & 6.14 & 5368  \\
		{} & 4 & 19.05 & 31.23 & 12.35\% & 60.19 & 6.72 & 6228 \\
		\hline
		\multirow{3}{*}{PEMSD8} & \bf{2} & \bf{14.92} & 24.17 & \bf{9.59}\% & \bf{32.43} & \bf{2.97} & \bf{2982} \\
		{} & 3 & 14.94 & \bf{24.14} & 9.61\% & 37.74 & 3.54 & 3410  \\
		{} & 4 & 14.96 & 24.18 & 9.83\% & 41.33 & 3.96 & 3740 \\
		\bottomrule[0.8pt]
	\end{tabular}
	}
\end{table}

\subsection{Hyperparameter Study}
To further evaluate the effects of hyperparameter tuning, we conducted a series of experiments on several core parameters on the PEMSD4 and PEMSD8 datasets.

{\noindent \bf Effect of Node Embedding Dimension $d$.}
To investigate the effect of the dimension of node embedding $E_{\phi}$ and temporal embeddings $E_1, E_2$ on model performance, we chose $d$ from $\{1, 2, \cdots, 10\}$. The results are shown in Figure \ref{fig_parar}(a) (red triangles indicate the best points). We observe that either too small (underfitting) or too large (overfitting) values of $d$ have different degrees of impact on the prediction performance.

{\noindent \bf Effect of Window Size $s$.}
The size of the cut over the time window of the input series for MS-GRU has a significant impact on model performance. The window size $s$ for grouping multiple time steps out of the $12$ time steps in the input data can be $2$, $3$, $4$, or $6$. The experimental results are shown in Fig. \ref{fig_parar}(b). The optimal $s$ is not necessarily the same for different data sets.

{\noindent \bf Effect of Convolution Kernel Number $K$.}
The results of the study on the number of convolution kernels $K \geq 2$ are shown in Table \ref{table6}, where we vary $K$ from $2$, $3$ to $4$.  From the experimental results, we can see that a larger convolution depth does not improve the prediction performance, but instead incurs longer training time and memory cost. Therefore, for our model, we set $K$ to $2$.

\begin{figure*}[!t]
\centering
\subfloat[]{\includegraphics[width=2.3in]{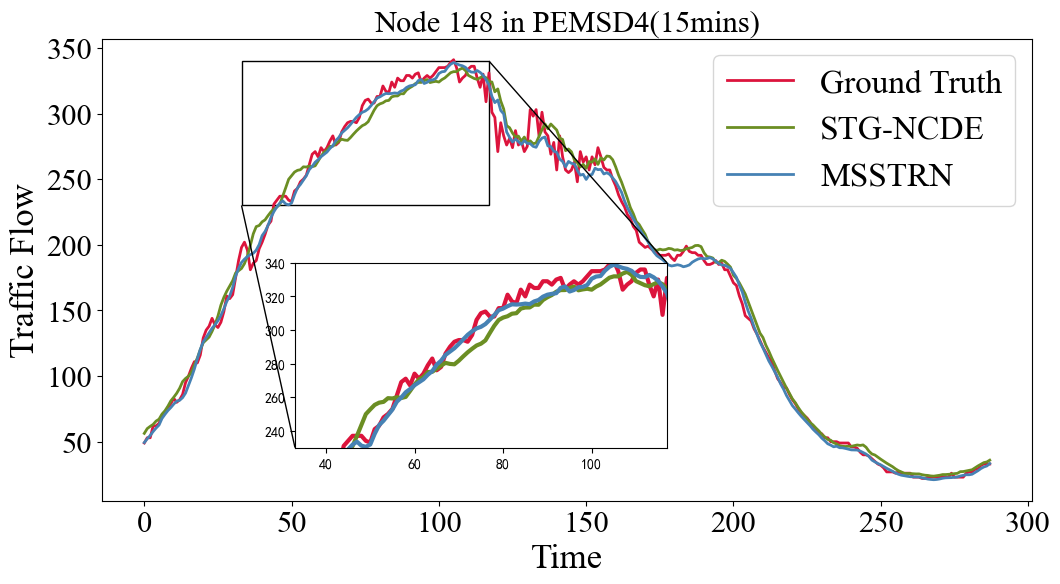}%
\label{fig_visual1}}
\hfil
\subfloat[]{\includegraphics[width=2.3in]{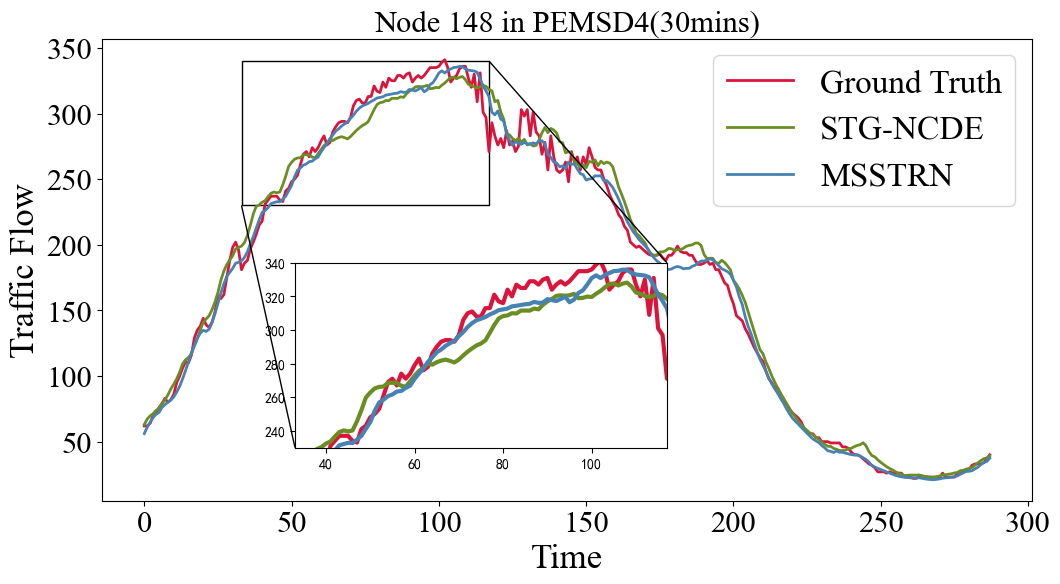}%
\label{fig_visual2}}
\hfil
\subfloat[]{\includegraphics[width=2.3in]{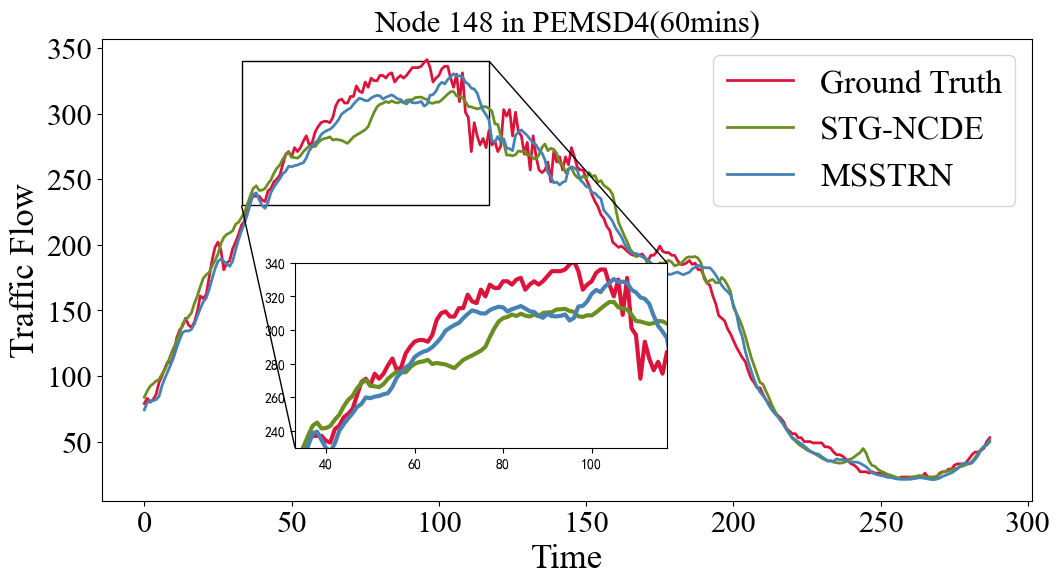}%
\label{fig_visual3}}

\subfloat[]{\includegraphics[width=2.3in]{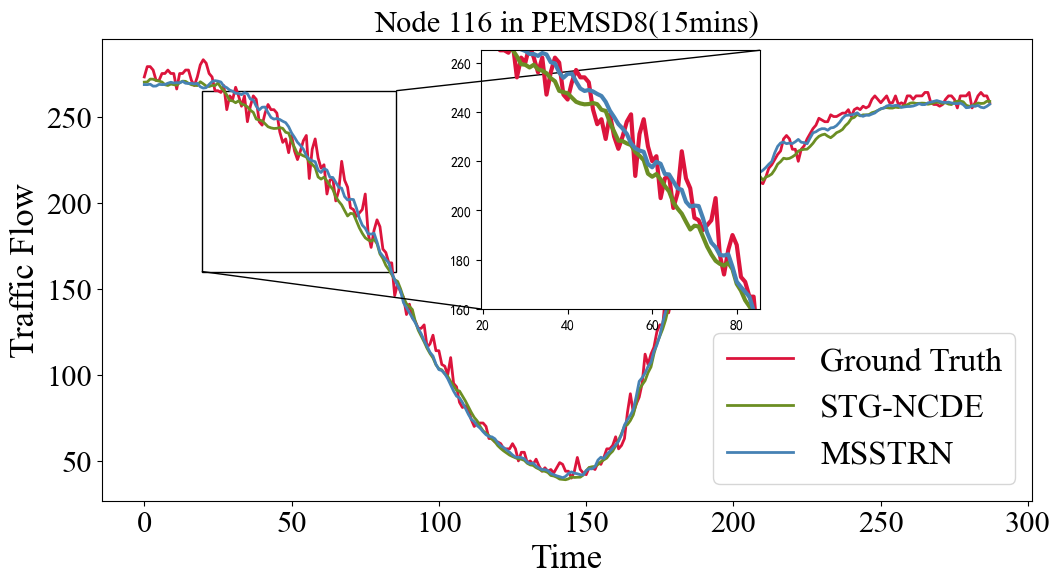}%
\label{fig_visual4}}
\hfil
\subfloat[]{\includegraphics[width=2.3in]{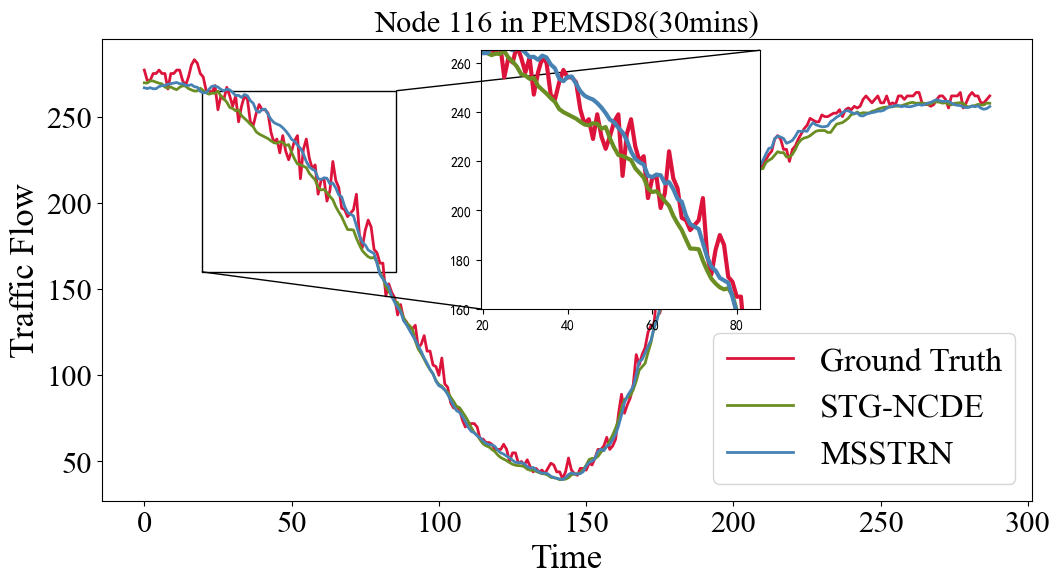}%
\label{fig_visual5}}
\hfil
\subfloat[]{\includegraphics[width=2.3in]{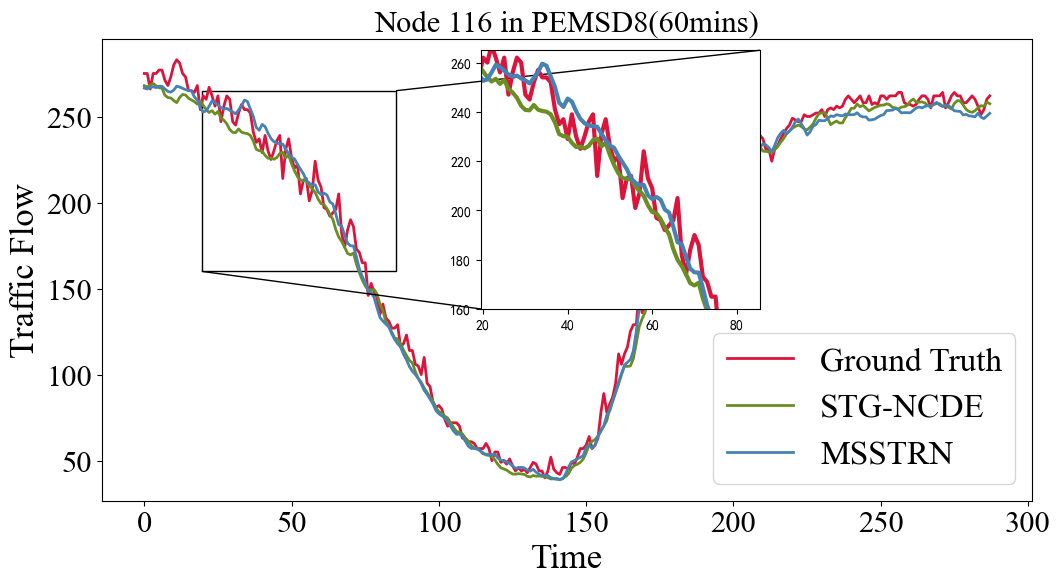}%
\label{fig_visual6}}
\caption{Traffic Flow Forecasting Visualization.}
\label{fig_visual}
\end{figure*}

To facilitate the replication of subsequent work, the hyperparameters for \emph{MSSTRN} to achieve optimal performance on each dataset are as follows:
\begin{itemize}
    \item \textbf{\emph{PEMSD3}}: the dataset batch size is $64$, the learning rate is $0.003$, the weight decay coefficient is $0.0002$, the number of convolution kernels $K=2$, the number of heads of STSAtt $h = 4$, the node embedding dimension is $7$ and the size of sub-windows $s=3$. 
    
    \item \textbf{\emph{PEMSD4}}: the dataset batch size is $64$, the learning rate is $0.003$, the weight decay coefficient is $0.0$, the number of convolution kernels $K=2$, the number of heads of STSAtt $h = 4$, the node embedding dimension is $9$ and the size of sub-windows $s=3$. 
    
    \item \textbf{\emph{PEMSD7}}: the dataset batch size is $64$, the learning rate is $0.003$, the weight decay coefficient is $0.0$, the number of convolution kernels $K=2$, the number of heads of STSAtt $h = 4$, the node embedding dimension is $9$ and the size of sub-windows $s=3$. 
    
    \item \textbf{\emph{PEMSD8}}: the dataset batch size is $64$, the learning rate is $0.003$, the weight decay coefficient is $0.0$, the number of convolution kernels $K=2$, the number of heads of STSAtt $h = 4$, the node embedding dimension is $6$ and the size of sub-windows $s=2$.  
\end{itemize}

\subsection{Flow Visualization}
We visualize the prediction results and ground truth of our model and STG-NCDE for PEMSD4 and PEMSD8 at $15$, $30$, and $60$ minutes ahead, as shown in Figure \ref{fig_visual}. Nodes $148$ and $116$ are two representative sensors on the PEMSD4 and PEMSD8 dataset, respectively. While the prediction curves of STG-NCDE at most time points are similar to ours, the highlighted sections in the box demonstrate the stronger prediction performance of our method in challenging situations (e.g., peak periods and high traffic fluctuations). In addition, the prediction curves of STG-NCDE in the long term differ more from the ground truth, and our model still maintains a higher accuracy.

\subsection{Model Building Study}
% \begin{table}[!t]
% 	\renewcommand{\arraystretch}{1.2}
% 	\centering
%  	\caption{Model building experiments on PEMSD8}
% 	\label{table7}
% 	\resizebox{.95\columnwidth}{!}{
% 	\begin{tabular}{c|c c c c c c}
% 		\toprule[0.8pt]
% 		{Model}& MAE & RMSE & MAPE & Training & Inference & Memory  \\
% 		\midrule
% 		{SS} & 15.44 & 24.53 & 10.06\% & 10.00 & 1.18 & 1587 \\
% 		{MS} & 17.97 & 28.45 & 11.45\% & 17.44 & 2.19 & 2697 \\
% 		{SS-SS} & 15.11 & 24.21 & 9.85\% & 20.64 & 2.39 & 2673 \\
% 		{MS-MS} & 18.61 & 29.33 & 11.61\% & 42.95 & 5.20 & 4885 \\
% 		{SS-MS} & 15.32 & 24.54 & 9.85\% & 32.23 & 3.93 & 4053 \\
% 		MS-SS (MSSTRN) & 15.05 & 24.16 & \bf{9.60}\% & 29.63 & 3.80 & 3659 \\
% 		{SS-SS-MS} & 15.01 & 24.16 & 9.61\% & 43.98 & 5.18 & 5147 \\
% 		{MS-SS-SS} & \bf{14.99} & \bf{24.05} & 9.68\% & 41.21 & 4.67 & 4669 \\
% 		{MS-MS-SS} & 15.46 & 24.60 & 9.94\% & 53.36 & 6.20 & 5725 \\
% 		{SS-SS-SS} & 15.43 & 24.52 & 10.15\% & 32.06 & 3.59 & 3673 \\
% 		\bottomrule[0.8pt]
% 	\end{tabular}
% 	}
% \end{table}

\begin{table*}[!t]
	\renewcommand{\arraystretch}{1.2}
	\centering
 	\caption{Model Building Experiments on PEMSD4 and PEMSD8}
	\label{table7}
	\resizebox{\linewidth}{!}{
	\begin{tabular}{c | c | c c c c c c c c c c | c}
		\toprule[0.8pt]
		{Datasets}& {Metrics} & SS & MS & SS-SS & MS-MS & SS-MS & SS-SS-SS & SS-SS-MS & SS-MS-MS & MS-MS-SS & MS-SS-SS & MS-SS(Ours) \\
		\midrule
		\multirow{6}{*}{PEMSD4} & {MAE} & 19.01 & 19.31 & 18.95 & 19.08 & 18.83 & 18.89 & 19.00 & 18.88 & 18.90 & 19.01 & \bf{18.77}\\
		{} & {RMSE} & 31.20 & 31.49 & 31.64 & 31.25 & 31.34 & 31.46 & 31.81 & 31.19 & 30.92 & 31.36 & \bf{30.91} \\
		{} & {MAPE} & 12.60\% & 12.60\% & 12.54\% & 12.33\% & 12.39\% & 12.28\% & 12.43\% & 12.36\% & 12.51\% & 12.32\% & \bf{12.14\%} \\
		{} & {Train.} & 12.15 & 20.28 & 32.04 & 53.84 & 44.21 & 52.81 & 68.70 & 79.36 & 72.32 & 60.06 & 40.02 \\
		{} & {Infer.} & 1.45 & 2.73 & 3.44 & 6.59 & 5.17 & 5.51 & 7.70 & 9.27 & 8.44 & 6.72 & 4.71 \\
		{} & {Mem.} & 2314 & 2868 & 4252 & 5902 & 5352 & 6032 & 7312 & 8264 & 7560 & 6532 & 4752 \\
		\hline
            \multirow{6}{*}{PEMSD8} & {MAE} & 15.21 & 15.57 & 15.12 & 15.02 & 15.14 & 15.13 & 14.91 & 15.24 & 15.04 & \bf{14.86} & 14.92 \\
		{} & {RMSE} & 24.42 & 24.65 & 24.53 & 24.19 & 24.45 & 24.53 & 24.29 & 24.51 & \bf{24.15} & 24.22 & 24.17 \\
		{} & {MAPE} & 9.86\% & 9.91\% & 9.70\% & 9.65\% & 9.57\% & 9.80\% & 9.50\% & 9.73\% & 9.58\% & \bf{9.46\%} & 9.59\% \\
		{} & {Train.} & 8.84 & 20.44 & 20.07 & 48.36 & 34.94 & 31.33 & 48.61 & 62.68 & 59.76 & 43.37 & 32.43 \\
		{} & {Infer.} & 1.09 & 1.69 & 2.32 & 4.07 & 3.34 & 3.51 & 4.74 & 5.71 & 5.25 & 4.13 & 2.97 \\
		{} & {Mem.} & 1592 & 1984 & 2682 & 3798 & 3432 & 3684 & 4406 & 5108 & 4672 & 3958 & 2982 \\
		\bottomrule[0.8pt]
	\end{tabular}
	}
\end{table*}

To explore the effect of stacking choice and the number of layers of SS-GRU and MS-GRU on model performance, we sampled different stacks to construct multiple prediction models, and then conducted experiments on datasets PEMSD4 and PEMSD8. For example, MS-SS is a two-layer spatial-temporal recurrent neural network model stacked with one layer of MS-GRU followed by one layer of SS-GRU, where SS denotes SS-GRU and MS denotes MS-GRU. If MS is executed after the SS operation (e.g., SS-MS), then Equation \ref{eq11} in the output layer uses $H_m^{(t-1)}$ as input. To facilitate the experiments, all the models are set to the same parameters.

The experimental results are shown in Table \ref{table7}. We can draw the following main conclusions: (1) The comparison of the results of SS with MS-SS or SS-SS with MS-SS-SS show that the prediction performance of the prediction models constructed by choosing only SS-GRU or MS-GRU is greatly reduced. (2) Performing the MS-GRU operation first seems to improve the prediction performance of the models. For example, MS-SS outperforms SS-MS and MS-SS-SS outperforms SS-SS-MS. (3) An increase in the number of layers, although it increases the training cost, may also improve the prediction performance, e.g. MS-SS-SS and MS-MS-SS on PEMSD8. Overall, both SS-GRU and MS-GRU can capture unique spatial-temporal dependencies, and a reasonable combination may construct high-performance prediction models. In this paper, we use MS-SS (named MSSTRN) for prediction experiments on all the datasets to ensure uniformity of prediction models.

\section{Conclusion}
In this paper, we design a new spatial-temporal prediction model named MSSTRN, which consists of two different recurrent neural networks: the Single Step Gate Recurrent Unit and the  Multi-Step Gate Recurrent Unit, which integrate adaptive position graph convolution and attention mechanisms into recurrent neural networks, respectively. Specifically, we propose a adaptive position graph generation module for spatial modeling. In addition, to achieve simultaneous capture of spatial-temporal dependencies, we propose a spatial-temporal synchronous attention mechanism. Extensive experimental results demonstrate the excellent predictive performance of our model. both APGG and STSAtt are generic modules and we believe they can be well applied to modeling other spatial-temporal tasks.

% \begin{figure}[!t]
% \centering
% \includegraphics[width=2.5in]{fig1}
% \caption{Simulation results for the network.}
% \label{fig_1}
% \end{figure}

% \begin{figure*}[!t]
% \centering
% \subfloat[]{\includegraphics[width=2.5in]{fig1}%
% \label{fig_first_case}}
% \hfil
% \subfloat[]{\includegraphics[width=2.5in]{fig1}%
% \label{fig_second_case}}
% \caption{Dae. Ad quatur autat ut porepel itemoles dolor autem fuga. Bus quia con nessunti as remo di quatus non perum que nimus. (a) Case I. (b) Case II.}
% \label{fig_sim}
% \end{figure*}

\bibliographystyle{IEEEtran}
\bibliography{tits}

% \vspace{40pt}
\begin{IEEEbiography}[{\includegraphics[width=1in,height=1.25in,clip,keepaspectratio]{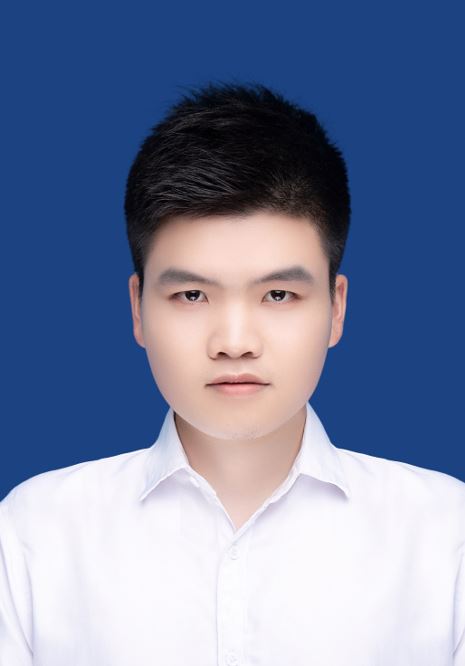}}]{Haiyang Liu}
is currently pursuing the master's degree with the Institute of Artificial Intelligence, Department of Computer Science and Technology, Soochow University, Suzhou, China. His research interests include spatial-temporal databases and intelligent transportation systems.
\end{IEEEbiography}
\begin{IEEEbiography}[{\includegraphics[width=1in,height=1.25in,clip,keepaspectratio]{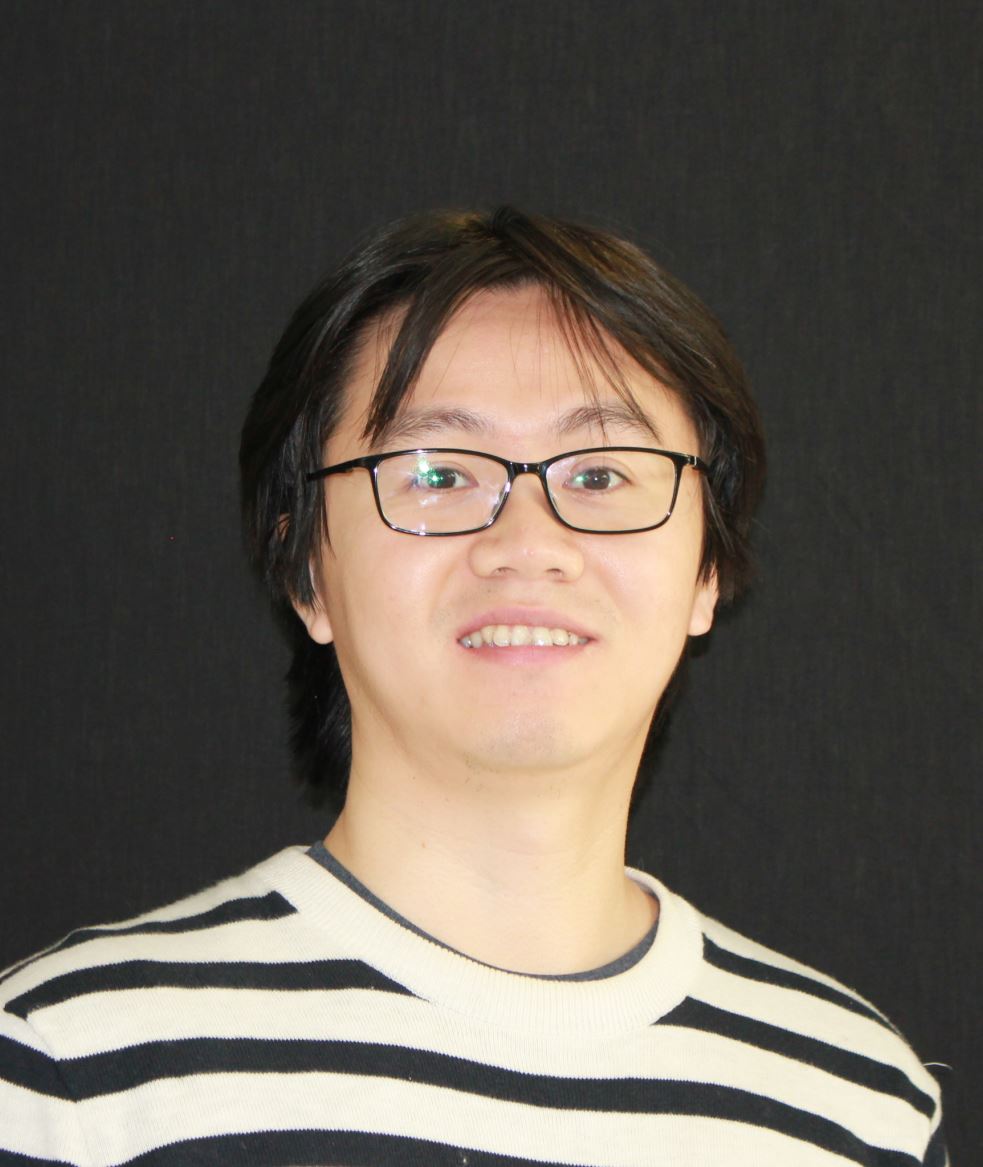}}]{Chunjiang Zhu}
 is an assistant professor in the Department of Computer Science at the University of North Carolina at Greensboro. He received his Ph.D. and Master in Computer Science from City University of Hong Kong and Chinese Academy of Sciences, respectively. His research interests include Machine Learning and Theory, Graph Algorithms, Chemoinformatics, and Cyber-Physical Systems.
\end{IEEEbiography}
\begin{IEEEbiography}[{\includegraphics[width=1in,height=1.25in,clip,keepaspectratio]{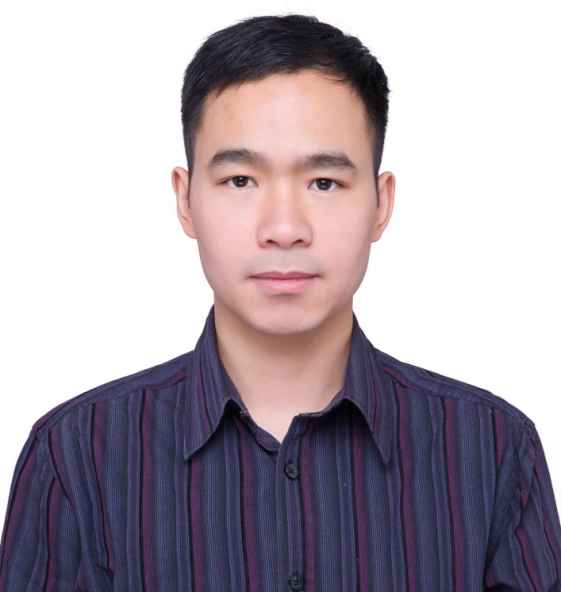}}]{Detian Zhang}
received the Ph.D. degree in computer science from City University of Hong Kong and University of Science and Technology of China in 2014. He is currently an associate professor with the Institute of Artificial Intelligence, School of Computer Science and Technology, Soochow University, Suzhou, China. His research interests include spatial-temporal databases and intelligent transportation systems.
\end{IEEEbiography}
\begin{IEEEbiography}[{\includegraphics[width=1in,height=1.25in,clip,keepaspectratio]{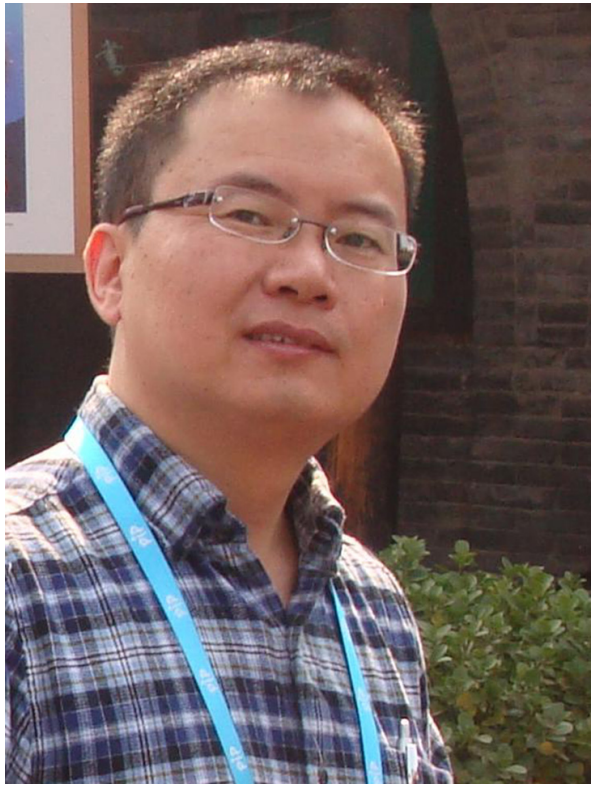}}]{Qing Li}
received the B.Eng. degree from Hunan University, Changsha, China, and the M.Sc. and Ph.D. degrees from the University of Southern California, Los Angeles, all in computer science. He is currently a Chair Professor (Data Science) and the Head of the Department of Computing, the Hong Kong Polytechnic University. He is a Fellow of IEEE, a Fellow of IET/IEE, a member of ACM-SIGMOD and IEEE Technical Committee on Data Engineering. He serves as a Steering Committee member of DASFAA, ER, ICWL, UMEDIA, and WISE Society. His current research interests include Multi-modal data management, Data warehousing and mining, Social media and Web services, and e-Learning Technologies.
\end{IEEEbiography}
\vfill

\end{document}